\documentclass[10pt,twocolumn,letterpaper]{article}

\usepackage{wacv}
\usepackage{times}
\usepackage{epsfig}
\usepackage{graphicx}
\usepackage{amsmath}
\usepackage{amssymb}

\usepackage{algorithm}
\usepackage{algorithmic}
\usepackage{amsmath,amssymb} 
\usepackage{color}
\usepackage{multirow} 
\usepackage{xspace}
\usepackage{capt-of}
\usepackage{enumitem}
\usepackage{calrsfs}
\DeclareMathAlphabet{\pazocal}{OMS}{zplm}{m}{n}

\usepackage{subcaption}
\usepackage{float}
\usepackage{enumitem}
\setlist{nolistsep}

\usepackage{array}
\newcolumntype{L}[1]{>{\raggedright\let\newline\\\arraybackslash\hspace{0pt}}m{#1}}
\newcolumntype{C}[1]{>{\centering\let\newline\\\arraybackslash\hspace{0pt}}m{#1}}
\newcolumntype{R}[1]{>{\raggedleft\let\newline\\\arraybackslash\hspace{0pt}}m{#1}}

\usepackage[pagebackref=true,breaklinks=true,letterpaper=true,colorlinks,bookmarks=false]{hyperref}
\usepackage{afterpage}

\wacvfinalcopy 

\def\wacvPaperID{157} 

\ifwacvfinal\fi
\begin{document}

\def\nofe{{NofE}\xspace}

\newcommand{\R}{\mathbb R}
\newcommand{\bx}{\mbox{\boldmath $x$}}

\def\expertconnect{{\scshape \footnotesize BranchConnect}\xspace}
\def\expertconnectbold{{\bf \scshape \footnotesize BranchConnect}\xspace}

\newcommand{\updates}[1]{{#1}}


\title{BranchConnect: Image Categorization\\with Learned Branch Connections}

\author{Karim Ahmed\hspace{2cm} Lorenzo Torresani \\
Department of Computer Science, Dartmouth College\\
{\tt\small karim@cs.dartmouth.edu, LT@dartmouth.edu}
}

%

\maketitle
\ifwacvfinal\thispagestyle{empty}\fi

\begin{abstract}
We introduce an architecture for image categorization that enables the end-to-end learning of separate visual features for the different classes to distinguish. The proposed model consists of a deep CNN shaped like a tree. The stem of the tree includes a sequence of convolutional layers common to all classes. The stem then splits into multiple branches implementing parallel feature extractors, which are ultimately connected to the final classification layer via learned gated connections. These learned gates determine for each individual class the subset of features to use. Such a scheme naturally encourages the learning of a heterogeneous set of specialized features through the separate branches and it allows each class to use the subset of features that are optimal for its recognition. We show the generality of our proposed method by reshaping several popular CNNs from the literature into our proposed architecture. Our experiments on the CIFAR100, CIFAR10, \updates{ImageNet}, and Synth datasets show that in each case our resulting model yields a substantial improvement in accuracy over the original CNN. Our empirical analysis also suggests that our scheme acts as a form of beneficial regularization improving generalization performance.
\end{abstract}


\section{Introduction}

In this paper we consider the problem of image categorization in scenarios involving a large number of classes. The recent introduction of several large-scale class-labeled image datasets~\cite{ImageNet, Krizhevsky:TR2009, Jaderberg14c, places} has stimulated active research on this topic. Deep convolutional neural networks (CNNs)~\cite{AlexNet, SimonyanZisserman:ICLR2015, HeEtAl:ICCV2015, GoogleLeNet, he2016deep} have emerged as highly effective models for visual recognition tasks. One of the reasons behind the success of deep networks in this setting is that they enable the learning of features that are explicitly optimized to discriminate as well as possible the given classes. 

It can be noted that the final fully-connected layer in a CNN can be viewed as implementing a separate linear classifier of the learned features for each of the classes to distinguish. But these class-specific classifiers operate all on the {\em same} feature representation, corresponding to the activations of the preceding layer. Under this model, the only obvious way to make the final classifiers more accurate is to either increase the number of features (e.g., by adding more neurons in the preceding layer) or make the features more expressive (e.g., by increasing the depth of the network). However, both these strategies increase the training cost and render the model more prone to overfitting. Furthermore, it can be argued that a single, jointly learned representation optimized on the overall problem of discriminating a large number of classes neglects to capture the uneven separability between classes, as some categories are inherently more difficult to distinguish than others or may require fine-grained distinction between similar classes.

To address these limitations, we propose a CNN architecture that makes it possible to learn ``dedicated'' or ``specialized'' features for each class but it still does so by means of a single optimization over all classes. This goal is realized through two mechanisms: a multi-branch network and learned gated connections between each class-specific classifier and the branches of the network. Let us consider each of these two mechanisms in turn. The multi-branch net has a tree-structure. It starts with a single stem corresponding to a sequence of convolutional layers common to all classes. The stem then splits into multiple branches implementing parallel feature extractors. Each branch computes a different representation, specialized for a subset of classes. In our scheme, the number of branches, $M$, is chosen to be smaller than the number of classes, $C$, (i.e., $M < C$) in order to parsimoniously share the features. Our model allows each class to use a subset of the $M$ features. The selection of features for each class is learned as part of the optimization by means of gated connections linking the $C$ neurons in the final fully-connected layer to the $M$ branches. The gate of a class effectively controls the subset of features used to recognize that category. Because each class can leverage a different set of features, the representation can be specialized to distinguish fine-grained categories that would be difficult to tell apart on the basis of a single global representation for all classes. At the same time, because all $M$ features originate from a common convolutional stem, the number of parameters in our model remains small compared to an ensemble of separate models. Furthermore, a single model allows us to obtain a diversified set of features by means of {\em one} optimization over the $C$ classes. The optimization learns jointly the weights in the network as well as the gates defining the subset of features for each class.

In summary, this work provides several contributions:
\begin{itemize}
\item We present a multi-branch, gated architecture that enables the learning of different features for the different classes in a large-scale categorization problem.
\item The learning is end-to-end and it is posed as optimization of a single learning objectives over all classes.
\item The training runtime is comparable to that of learning a traditional single-column network with the same depth and number of parameters.
\item We demonstrate the generality of our approach by reporting results on \updates{4} different datasets. In each case, our adaptation of traditional CNNs into \expertconnect yields a substantial improvement in accuracy.
\item We perform a sensitivity analysis of the effect of varying the number of branches and the number of branch features used by each class.
\item We present an empirical study suggesting that our proposed scheme acts as a beneficial regularizer, yielding larger training error but improved test performance compared to a flat architecture of the same depth without branches and gated connections.
\end{itemize}


\section{Related Work}

Our approach bears closest similarity with deep learning methods that learn models or representations specialized for the different categories in a classification problem. Hinton et al.~\cite{HintonEtAl:arXivDISTILL} achieve this goal by means of an ensemble of deep networks consisting of one full model trained to discriminate all classes (the generalist) and many specialists that improve the recognition of fine-grained classes confused by the full model. The approach of Farley et al.~\cite{FarleyEtAl:ICLR2015} explores a similar idea but with the difference that the specialists are trained on top of the low-level features already learned by the generalist, thus giving rise to a unified model. Yan et al.~\cite{HD-CNNICCV15} and Ahmed et al.~\cite{AhmedEtAl:ECCV16} propose a hierarchical decomposition of the categories where learned clusters of categories are first distinguished by a coarse-category classifier and then fine-grained classes within each cluster are discriminated by an expert. All these prior approaches~\cite{HintonEtAl:arXivDISTILL, FarleyEtAl:ICLR2015, HD-CNNICCV15, AhmedEtAl:ECCV16} differ from ours in the fact that they decompose the training into multiple stages, where first a full generalist is trained over all classes, and then separate expert networks or subnetworks are learned for different subsets of categories. Instead, our method learns class-specific representations by optimizing a single learning objective over all classes and does not require costly pretraining of a generalist. This enables fast training of our model, with computational cost  equivalent to that of training a traditional single-column CNN of the same size. In the experimental section we compare our approach to HD-CNN~\cite{HD-CNNICCV15} and Network of Experts~\cite{AhmedEtAl:ECCV16} and demonstrate that, for the same depth and number of parameters, our model yields significantly higher accuracy. Unfortunately, we cannot directly compare with the methods described in~\cite{HintonEtAl:arXivDISTILL, FarleyEtAl:ICLR2015} as their models were trained on an internal Google dataset involving 100M labeled images. Neither the models, nor the softwares, nor the dataset have been publicly released. 

Aljundi et al.~\cite{AljundiExpertGate} propose the use of autoencoders to learn a representation that is specialized for each different task in a lifelong-learning setting, where new tasks are added to the model sequentially giving rise to an ensemble of experts. This work also adopts a gating mechanism. But the gates are used to route the input example to the most suitable expert model. Instead, our approach is designed to operate in a scenario where all the data is fully available at the time when training is started, and it involves the use of gated connections to diversify the features used by different classes rather than to route the computation to separate models. 

Our model also shares similarities with the mechanism of skipped connections popularized by fully convolutional networks~\cite{FCNDarrell} and used also in residual learning~\cite{he2016deep}. Skipped connections enable the additive fusion of activations produced by different layers in the network. However, skipped connections define a priori the layers to add together in the architecture. Instead, our approach learns the branch layers to combine via gated connections that are trained simultaneously with the weights of the network. 

Our approach is also related to dropout~\cite{dropout}, which has been show to act as a regularizer preventing overfitting. In our experiments we demonstrate that our \expertconnect also produces a regularization effect. However, while in dropout connections are randomly dropped during each training iteration, our method explicitly learns the connections that are most beneficial to eliminate and those that should be preserved. Stochastic pooling~\cite{ZeilerFergus:2013} is another regularization mechanism leveraging stochasticity during learning. Our gating function is closely related to stochastic pooling, as both rely on sampling a multinomial distribution during training. However, while in stochastic pooling the multinomial is defined from activations in a region, in our case the parameters of the multinomial are learned via backpropagation and do not change with the input. Furthermore, while stochastic pooling is used to reduce the dimensionality of activation volumes in deep CNNs, the gated connections of \expertconnect are designed to differentiate the features for each category in a classification problem.

Finally, our work falls in the broad genre of multi-branch networks, which have been adopted in previous work to address a wide array of tasks ranging from combining decision trees with convolutional networks~\cite{KontschiederFCB16},  to autoencoding~\cite{IrsoyA15} as well as feature pooling~\cite{LeeGT16}. 




\section{Technical Approach}
\label{sec:approach}

\begin{figure}[!t]
\begin{center}
\centerline{\includegraphics[width=\linewidth]{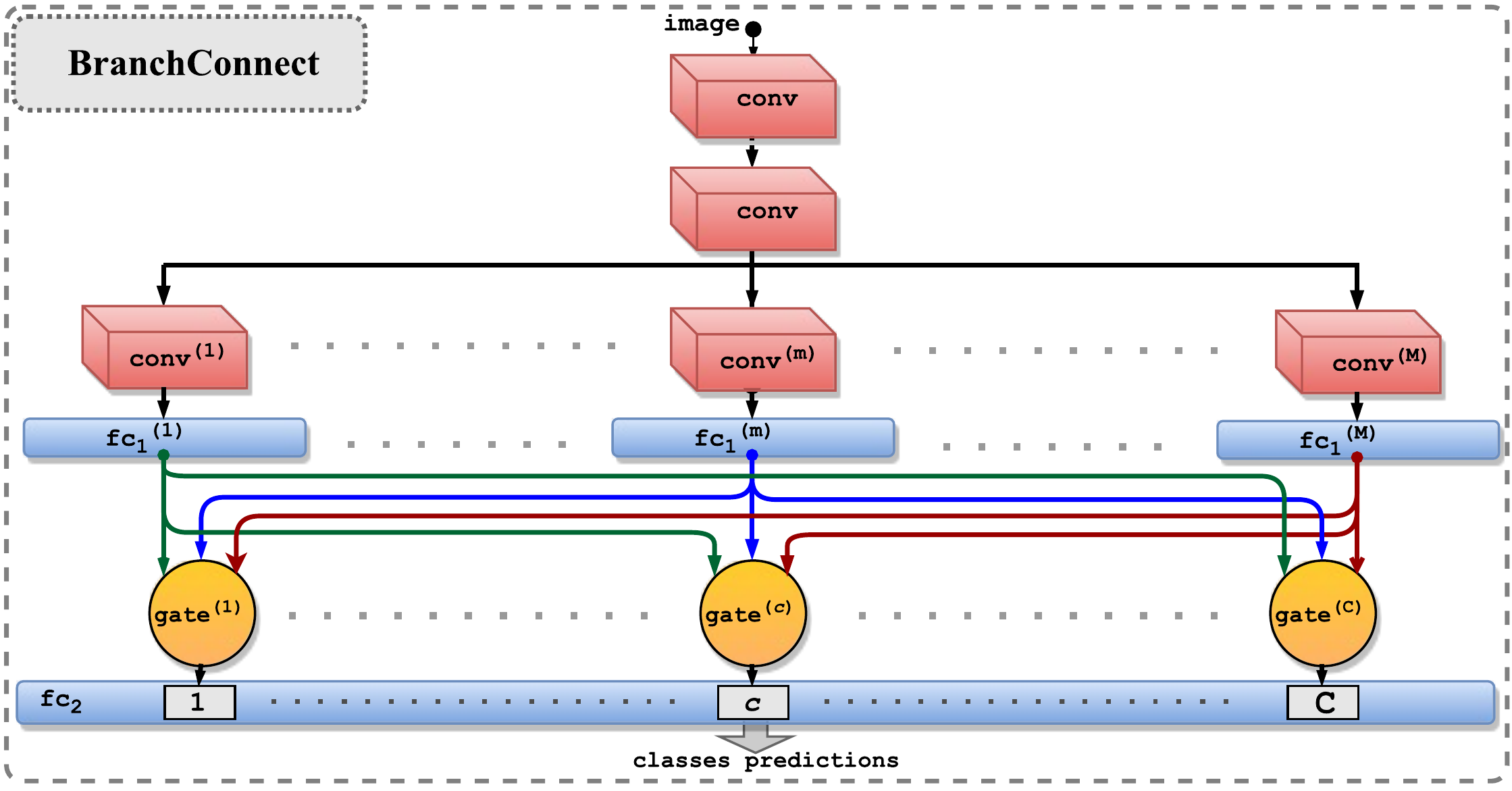}}  
\end{center}
\caption{The architecture of \expertconnect for classification of $C$ classes. The branches implement $M<C$ parallel feature extractors. Class-specific gates connect the $M$ branches to the $C$ classes in the last fully-connected layer.}
\label{gnofe_model}

\end{figure}


In this section we present our proposed technical approach. We begin by introducing the notation that will be used throughout this paper. We assume we are given a training dataset $\mathcal{D}$ of  $N$ class-labeled images: $\mathcal{D} = \{(x^1,y^1), \hdots, (x^N,y^N)\}$ where $x^i$ represents the $i$-th RGB image and $y^i \in \{1, \hdots, C\}$ denotes its associated class label, with $C$ indicating the total number of classes. 

In subsection~\ref{sec:buildgnofe} we describe the architecture of \expertconnect. In subsection~\ref{sec:learngnofe} and subsection~\ref{sec:inferencenofe} we discuss the training and inference procedures, respectively.

\subsection{The architecture of {\bf \expertconnect}}
\label{sec:buildgnofe}

The architecture of \expertconnect is a tree-structured network, as illustrated in Figure~\ref{gnofe_model}. It consists of a {\em stem} that splits into $M < C$ branches, where $M$ is a hyper-parameter that controls the complexity of our model. The stem consists of a sequence of convolutional layers possibly interleaved by pooling layers. Each branch contains one or more convolutional/pooling layers, followed by zero, one or more fully-connected layers (in our experiments we present results for a variety of models). The branches have identical architecture but different parameters. The \expertconnect model culminates into a fully-connected layer of $C$ neurons using the softmax activation function to define a proper distribution over the $C$ classes to discriminate. This last layer takes as input the activations from the $M$ branches and it is equivalent in role to the last fully-connected layer of a traditional CNN for categorization. However, in \expertconnect each of these $C$ neurons has a dedicated {\em branch gate} that controls the input effectively fed to the neuron. More specifically, let us consider the $c$-th neuron in the last fully-connected layer of $C$ units. We refer to this neuron as the {\em neuron classifier} of class $c$, since it is responsible for computing the probability that the input image belongs to class $c$. The branch gate of this class is a {\em learned} binary vector ${\bf g}^{b}_{c} = \left[{g}^{b}_{c,1}, {g}^{b}_{c,2}, \hdots, {g}^{b}_{c,M}\right]^\top \in \{0, 1\}^M$ specifying the branches taken into consideration by the neuron classifier to predict the probability of class $c$. If ${g}^{b}_{c,m} = 1$, then the activation volume produced by the $m$-th branch is fed as input to the neuron of class $c$. If ${g}^{b}_{c,m} = 0$, then the computation from the $m$-th branch is ignored by the classifier for class $c$. Thus, if we denote with $E_m$ the output activation tensor computed by the last layer of the $m$-th branch, the input $F_c$ to the $c$-th neuron will be given by the following equation:
\vspace{-0.4cm}
\begin{equation} 
F_c = \sum_{m=1}^{M} g^{b}_{c,m} \cdot E_{m}
\label{eq:gateoutput}
\end{equation}
The interpretation is that the branch gate ${\bf g}^{b}_{c}$ adds {\em selectively} the information from the $M$ branches by choosing the branches that are most salient for the classification of class $c$. Under this scheme, each branch can therefore specialize to compute features that are relevant only to a subset of the classes. We also point out that depending on the constraints posed over ${\bf g}^{b}_{c}$, different interesting models can be realized. For example, by introducing the constraint that $\sum_m {g}^{b}_{c,m} = 1$, only one branch will be {\em active} for each neuron $c$ (since ${g}^{b}_{c,m}$ must be either 0 or 1). Such a model would effectively partition
the set of $C$ classes into $M$ disjoint clusters, where branch $m$ is trained to discriminate among the classes in cluster $m$. It can be noted that at the other end of the spectrum, if we set ${g}^{b}_{c,m}=1$ for all branches $m$ and classes $c$, then all classifiers in the last layer would be operating on the same input. In our experiments we will demonstrate that the best results are achieved for a middle ground between these two extremes, i.e., by connecting each neuron classifier to exactly $K$ branches where $K$ is a cross-validated hyper-parameter such that $1< K < M$. As discussed in the next section, the gate ${\bf g}^{b}_{c}$ of each class $c$ is learned simultaneously with all the other weights in the network via backpropagation. 

We point out that Eq.~\ref{eq:gateoutput} uses {\em additive} selective fusion of the output produced by the branches. We have also tried stacking (rather than adding) the feature maps of the active branches but this increases by a multiplicative factor the parameters in the last layer and in our experiments this approach produced results inferior to the additive scheme.

\subsection{Training {\bf \expertconnectbold}}
\label{sec:learngnofe}

The training of our model is end-to-end and it is done by optimizing via backpropagation a given learning objective $\ell$ over the $C$ classes of dataset $\mathcal{D}$. However, in the case of \expertconnect, the objective is optimized with respect to not only the weights of the network but also the branch gates, which are viewed as additional parameters in the model.

In \expertconnect, the weights of the convolutional and fully connected layers are real values, as in traditional CNNs. Instead, the branch gates are binary, which render optimization more challenging. To learn these binary parameters, we adopt a procedure inspired by the  algorithm proposed in~\cite{NIPS2015_5647} to train neural networks with binary weights. During training we store and update a real-valued version ${\bf g}^{r}_c \in \left[0,1\right]^M$ of the branch gates, with entries clipped to lie in the continuous interval from 0 to 1. 

In general, the training of a CNN consists of three steps: 1) forward propagation, 2) backward propagation, and 3) parameters update. We stochastically binarize the real-valued branch gates into binary-valued vectors ${\bf g}^{b}_c \in \{0,1\}^M$ only during  the forward propagation and backward propagation (steps 1 and 2), whereas during the parameters update (step 3), the method updates the real-valued branch gates ${\bf g}^{r}_c$. The remaining weights of the convolutional and fully connected layers are optimized using standard backpropagation. In the next subsections we discuss in further detail the gate training procedure, under the assumption that at any time there can be only $K$ active entries in the binary branch gate ${\bf g}^{b}_c$, where $K$ is a predefined integer hyper-parameter with $1\leq K \leq M$. In other words, we impose the following constraints: 
\vspace{-0.3cm}
\begin{eqnarray}
\sum_{m=1}^M g^{b}_{c,m} = K, \forall c\in\{1,\hdots,C\} \nonumber\\
g^{b}_{c,m} \in \{0,1\}, \forall c\in\{1,\hdots,C\} \text{ and } \forall m\in\{1,\hdots,M\}.\nonumber
\end{eqnarray}
These constraints imply that each classifier neuron in the last layer receives input from exactly $K$ branches. The entire training procedure for the branch gates is summarized in Algorithm~\ref{alg:trainalg1} and discussed in detail below. 



\subsubsection*{\bf Branch Gates: Forward Propagation}

During the forward propagation, our algorithm first normalizes the current $M$ real-valued branch gates $g^{r}_{c,m}$ for each class $c$ to sum up to 1. This is done so that $\text{Mult}(g^r_{c,1},g^r_{c,2},\hdots,g^r_{c,M})$ defines a proper multinomial distribution over the $M$ branch connections of the $c$-th neuron classifier. Then, the binary branch gate ${\bf g}^{b}_{c}$ is stochastically generated by drawing $K$ {\em distinct} samples $i_1,i_2,\hdots,i_K \in \{1, \hdots, M\}$ from the multinomial distribution over the branch connections. 
Then, the entries corresponding to the $K$ samples are activated in the binary branch gate vector, i.e., $g^{b}_{c,i_k} \leftarrow 1$, for $k=1,...,K$. The input activation volume to the neuron classifier for each class $c$ is then computed according to Eq.~\ref{eq:gateoutput} from the sampled binary branch gates and the final prediction is obtained. 

We have also experimented with a deterministic procedure that sets the active branch connections in ${\bf g}^{b}_{c}$ to correspond to the $K$ largest values in ${\bf g}^{r}_{c}$. However, we found that this often causes the binary gate vector ${\bf g}^{b}_{c}$ to remain stuck at the initial configuration. As also reported in~\cite{NIPS2015_5647}, we found the stochastic assignment of binary gates according to the real-valued probabilities to yield much better performance. In all our experiments we initialize the real-valued scalar branch gates $g^{r}_{c,m}$ to $0.5$. This allows the training procedure to explore different connections in the first few iterations.


\subsubsection*{\bf Branch Gates: Backward Propagation}
In the backward propagation step, our method first computes the gradient of the mini-batch loss with respect to the input volume of each neuron classifier, i.e., $\frac{\partial \pazocal{\ell}}{\partial F_{c}}$. Then, the gradient $\frac{\partial \pazocal{\ell}}{\partial E_{m}}$ with respect to each branch output is obtained via back-propagation from $\frac{\partial \pazocal{\ell}}{\partial F_{c}}$ and the current binary branch gates $g^{b}_{c,m}$.

\subsubsection*{\bf Branch Gates: Parameters Update}
As shown in Algorithm~\ref{alg:trainalg1}, in the parameter update step our algorithm computes the gradient with respect to the binary branch gates for each branch. Then, using these computed gradients and the given learning rate, it updates the real-valued branch gates via gradient descent. At this time we clip the updated real-valued branch gates to constrain them to remain within the valid interval $[0, 1]$. The same clipping strategy was adopted for the binary weights in the work of Courbariaux et al.~\cite{NIPS2015_5647}. 

\begin{algorithm}[ht!]
   \caption{Training Branch Gates with \expertconnect. }
\label{alg:trainalg1}

\begin{algorithmic}
{\footnotesize

   \STATE {\bfseries Input:} a minibatch of labeled examples $(x^i, y^i)$, $C$: number of classes, $M$: number of  branches, $K$: the number of active branch connections per class, $\eta$: learning rate,  $\pazocal{\ell}$: the loss over the minibatch, ${\bf g}^{r}_c \in \left[0,1\right]^M$: real-valued branch gates from previous training iteration.
   \STATE {\bfseries Output:} updated ${\bf g}^{r}_c,$ for all classes $c=1,\hdots, C$ 
   
   \STATE {\bfseries 1. Forward Propagation:}
   
    \FOR{$c \leftarrow  1$ {\bfseries to} $C$}
   
    \STATE Normalize the real-valued branch gates of class $c$ to sum up to 1:  $g^r_{c,m} \leftarrow \frac{g^r_{c,m}}{\sum_{m'=1}^{M} g^r_{c,m'}},$ for $m=1,\hdots, M$
   \STATE Reset binary branch gates: ${\bf g}^{b}_{c} \leftarrow {\bf 0}$   
   \STATE Draw $K$ {\em distinct} samples from multinomial branch gate distribution: $ i_1,i_2,\hdots,i_K \leftarrow  \text{Mult}(g^r_{c,1},g^r_{c,2},\hdots,g^r_{c,M})$ 
   \STATE Set active binary branch gates based on drawn samples: \\$g^{b}_{c,i_k} \leftarrow 1 \text{ for } k=1,...,K$
   \STATE Compute input $F_{c}$ to the $c$-th neuron, given branch activations $E_{m}$:  $F_{c} \leftarrow  \sum_{m=1}^{M} g_{c,m}^{b} \cdot E_{m}$ 
   \ENDFOR

   
   \STATE {\bfseries 2. Backward Propagation:} 
   
    \FOR{$c \leftarrow  1$ {\bfseries to} $C$}
  
   \STATE Compute $\frac{\partial \pazocal{\ell}}{\partial F_{c}}$ from $\ell$ and neuron classifier parameters
  
   \STATE Compute $\frac{\partial \pazocal{\ell}}{\partial E_{m}}$ given  $\frac{\partial \pazocal{\ell}}{\partial F_{c}}$, $g^{b}_{c,m}$ for $m=1,...,M$
   
   \ENDFOR

   
   \STATE {\bfseries 2. Parameter Update:} 
   
   \FOR{$c \leftarrow  1$ {\bfseries to} $C$}

   \STATE Compute $\frac{\partial \pazocal{\ell}}{\partial g^{b}_{c,m}}$  given  $\frac{\partial \pazocal{\ell}}{\partial F_{c}}$  and $E_{m}$, for $m=1,...,M$
   
\STATE $g^{r}_{c,m} \leftarrow $ \text{clip}($g^{r}_{c,m} - \eta \cdot \frac{\partial \pazocal{\ell}}{\partial g^{b}_{c,m}} )$ for $m=1,...,M$

   \ENDFOR

 }

\end{algorithmic}


\end{algorithm}

\subsection{Inference with {\bf \expertconnectbold}}
\label{sec:inferencenofe}

In order to perform test-time inference on new samples given a trained model with real-valued branch gates ${\bf g}^{r}_{c}$, we adopt a deterministic strategy, rather than the stochastic approach used during training. We simply set to 1  the entries of ${\bf g}^{b}_{c}$  that correspond to the largest $K$ values of ${\bf g}^{r}_{c}$, and leave all other entries set to 0. 

We have also experimented with using the non-binary gates ${\bf g}^{r}_{c}$ for inference at test time but found this approach to yield much lower performance. This is understandable given that the learning objective is computed and minimized using binary rather real-valued gates.





\section{Experiments}

We demonstrate the effectiveness and the generality of our  approach by presenting experiments using  \expertconnect models built from different CNN architectures on four different datasets: CIFAR-100~\cite{Krizhevsky:TR2009}, CIFAR-10~\cite{Krizhevsky:TR2009}, \updates{ImageNet}~\cite{ImageNet}, and the Synthetic Word Dataset (Synth)~\cite{Jaderberg14c,Jaderberg14d}.

\subsection{Reshaping a traditional CNN into a multi-branch net with \expertconnect}

In order to show the benefits of \expertconnect we present results obtained by reshaping several traditional CNNs from the literature into the form of our multi-branch architecture. We refer to the original architectures as the {\em base} models. Note that our approach requires only the {\em specification} of the base CNN architecture, i.e., no pre-trained parameters are needed. 

We evaluate a simple, single recipe to reshape each base model into an \expertconnect network. Let $P_c$ be the total number of convolutional/pooling layers of the base model, and $P_f$ the number of fully-connected layers following the convolutional/pooling layers. The stem of \expertconnect is formed by using  $P_c-1$ convolutional/pooling layers identical in specifications to the first $P_c-1$ layers of the base model. Then, we place in each branch the remaining convolutional layer of the base model followed by $P_f-1$ fully connected layers identical in specifications to the first $P_f-1$ fully connected layers of the base models (i.e, all fc layers except the last one). The $M$ branches have identical architecture but distinct parameters. Then, we place a final fully connected layer of size $C$ at the top. This layer is shared among all branches and is responsible for the final prediction. Neuron $c$ in this layer is connected through learned gate ${\bf g}^{b}_{c}$ to the last layer of the $M$ branches (see Figure~\ref{gnofe_model}).


\subsection{CIFAR-100}
CIFAR-100 is a dataset of 32x32 color images spanning $C=100$ classes. The training set contains 50,000 examples and the test set includes 10,000 images. We use this dataset to conduct a comprehensive study using different network architectures and settings. 

\subsubsection{Accuracy Gain for Different Architectures}

We begin by showing that \expertconnect yields consistent improvements irrespective of the specific architecture. To demonstrate this, we take five distinct architectures from prior work~\cite{he2016deep, HD-CNNICCV15, AhmedEtAl:ECCV16} and reshape them as \expertconnect. 

The architectures are listed below (full specifications are listed in Appendix~\ref{appendixA}). For this preliminary set of experiments we fix the number of \expertconnect branches $M$ to 10. For each architecture, we train 10 separate models for values of $K$ (the number of active branches per class) ranging from 1 to 10. As already discussed in section~\ref{sec:approach}, we build the \expertconnect network from each base model by placing all convolutional layers except the last one in the stem. Each branch then contains one convolutional layer (identical in specifications to the last convolutional layer of the base model) followed by fully-connected layers (identical to those in the base model), except for the last one. The last fully-connected layer is shared among all branches. (see Figure~\ref{gnofe_model}). 

\afterpage{
\begin{table}[t!]
\caption[Caption for LOF]{Classification accuracy (\%) \updates{(single crop)} on CIFAR-100 for 5 base architectures. \expertconnect uses $M=10$ branches. \textsc{G:}$K$/$M$ means that each gate has $K$ active connections. We report performance for $K=1$ and when choosing the best value of $K$.\footnotemark}
\small{
\begin{center}\
\setlength{\tabcolsep}{4pt} 
\renewcommand{\arraystretch}{1}
\label{table:results_cifar100}

\begin{tabular}{|p{0.2cm}||l|C{0.8cm}|C{1.2cm}|C{1.35cm}|}

\hline
 & Method & depth & \#params & Accuracy  \\
\hline\hline

\multirow{5}{*}{\rotatebox[origin=c]{90}{\bf \footnotesize  AlexNet-Quick}} &
Base Model V1 & 5 & 0.15M  & 44.3\\
& Base Model V2 & 5 & 1.20M & 40.26\\
\cline{2-5}
& \nofe  \cite{AhmedEtAl:ECCV16} & 6 & 1.27M & 49.09\\
\cline{2-5}


& \expertconnect G:1/10 & 5 & 1.20M & \textbf{53.28}\\
& \expertconnect G:5/10 & 5 & 1.20M & \textbf{54.62}\\

\hline\hline

\multirow{5}{*}{\rotatebox[origin=c]{90}{\bf \footnotesize Alexnet-Full}} &
Base Model V1 & 4& 0.18M & 54.04  \\
& Base Model V2 & 4& 0.64M & 50.42   \\
\cline{2-5}
&\nofe \cite{AhmedEtAl:ECCV16} & 5& 1.12M & 56.24  \\
\cline{2-5}

& \expertconnect G:1/10 & 4 & 0.64M & \textbf{57.34} \\
& \expertconnect G:6/10 & 4 & 0.64M & \textbf{60.27} \\

\hline\hline

\multirow{5}{*}{\rotatebox[origin=c]{90}{\bf \footnotesize NIN~\cite{NIN}}} &
Base Model V1 & 9 & 1.38M & 64.73  \\
&Base Model V2 & 9 & 1.61M & 65.24  \\
\cline{2-5}
&HD-CNN \cite{HD-CNNICCV15}   & n/a & n/a & 65.64  \\
\cline{2-5}
&\nofe \cite{AhmedEtAl:ECCV16}  & 11 & 4.66M & 65.91  \\
\cline{2-5}

&\expertconnect G:1/10 & 9 & 1.61M  & \textbf{66.10} \\
&\expertconnect G:5/10 & 9 & 1.61M  & \textbf{66.45} \\
\hline\hline

\multirow{4}{*}{\rotatebox[origin=c]{90}{\parbox{1.5cm}{\bf \footnotesize ResNet56~\cite{he2016deep}}}} &
Base Model V1    & 56 &  0.86M & 69.66  \\
&Base Model V2    & 56 &  1.47M  & 70.72   \\
\cline{2-5}
&\expertconnect G:1/10 & 56 &  1.47M  & \textbf{71.24} \\
&\expertconnect G:5/10 & 56 & 1.47M  &\textbf{71.98} \\
\hline\hline

\multirow{5}{*}{\rotatebox[origin=c]{90}{\bf \footnotesize ResNet56-4X~\cite{AhmedEtAl:ECCV16}}} &
Base Model V1~\cite{AhmedEtAl:ECCV16}  & 56 & 13.6M & 72.23  \\
&Base Model V2    &  56 &   25.4M  & 73.12   \\
\cline{2-5}
&\nofe  \cite{AhmedEtAl:ECCV16}& 58 &  25.5M & 74.71 \\
\cline{2-5}
&\expertconnect G:1/10 & 56 & 25.4M  & \textbf{75.55} \\
&\expertconnect G:5/10 & 56 &  25.4M &\textbf{75.72} \\
\hline

\end{tabular}
\end{center}
\vspace{-0.9cm}
}
\end{table}
\footnotetext{Note that some of the accuracies for \nofe and Base Models listed here differ slightly from those reported in~\cite{AhmedEtAl:ECCV16}. The differences are merely due to the fact that in~\cite{AhmedEtAl:ECCV16} some of the architectures were tested using multiple image crops, while our evaluation uses a single crop for all architectures.}
}

Here are the five base models for this experiment:

\noindent 1) {\em AlexNet-Quick.} This is a slightly modified version of the AlexNet model~\cite{AlexNet} adapted by Ahmed et al~\cite{AhmedEtAl:ECCV16} to work on the 32x32 images of CIFAR-100.  It consists of 3 convolutional layers and 2 fully-connected layers.  Thus, our \expertconnect net constructed from {\em AlexNet-Quick} includes two convolutional layers in the stem, while each branch contains one convolutional layer with the same specification as the third convolutional layer in the base model and one fully-connected layer.  

\noindent 2)  {\em AlexNet-Full.} This model is also taken from~\cite{AhmedEtAl:ECCV16}. This base CNN is slightly different from {\em AlexNet-Quick} as it has only one fully-connected layer instead of two layers, and it uses local response normalization layers. The accuracy of this base model is higher than {\em AlexNet-Quick}. The corresponding \expertconnect model consists of a stem that contains the first two convolutional layers. Each branch consists of only one convolutional layer with the same specification as the third convolutional layer in the base model. 

\noindent 3) {\em NIN.} This model is  a ``Network In Network" (NIN)~\cite{NIN}. NIN models do not use fully-connected layers. Instead, they employ as final layer a convolutional layer where the number of filters is equal to the number of classes ($C$). The final prediction is obtained by performing global average pooling over the feature maps of this layer. Thus, we build \expertconnect by placing in the stem all convolutional layers, except the last two. Each branch then contains only one convolutional layer. Finally the branches are connected via our binary gates to the final convolutional layer of $C$ filters followed by global average pooling for the final prediction.

\noindent 4) {\em ResNet56.} This is the 56-layer residual network originally described in~\cite{he2016deep}. In the \expertconnect model, the stem contains all the residual blocks except the last block which is included in the branches. The final fully-connected layer is shared among all branches.

\noindent 5) {\em ResNet56-4X.} This model is identical in structure to {\em ResNet56} but it uses 4 times as many filters in each convolutional layer and was shown in~\cite{AhmedEtAl:ECCV16} to yield higher accuracy on CIFAR-100.

The results achieved with these 5 architectures are shown in Table~\ref{table:results_cifar100}. For each architecture, we report the accuracy of the base model {\em ``Base Model V1"}  as well as that obtained with our \expertconnect. We report two numbers for \expertconnect: the first using only {\em one} active branch per class (i.e., $K=1$), the second obtained by choosing the best value of $K$ (ranging from 1 to 10) for each architecture. We also include for comparison the performance achieved by Network of Experts (NofE)~\cite{AhmedEtAl:ECCV16}, which also builds a branched architecture from the given base model. However, NofE does so by performing hierarchical decomposition of the classes using two separate training stages and it connects each class to only one branch by construction. For the case of NIN, we also include the performance reported in~\cite{HD-CNNICCV15} for the hierarchical HD-CNN built from this base model.

We can see from Table~\ref{table:results_cifar100} that \expertconnect outperforms the base model {\em ``Base Model V1"} for all five architectures. \expertconnect does also considerably better than NofE~\cite{AhmedEtAl:ECCV16} and HD-CNN~\cite{HD-CNNICCV15}, which are the most closely related approaches to our own. For all architecture the peak performance of \expertconnect is achieved when setting the number of active branches ($K$) to be greater than 1 (the best accuracy is achieved with $K=6$ for {\em AlexNet-Full} and with $K=5$ for the other four models).

It can be noted that \expertconnect involves more parameters than the base models {\em ``Base Model V1"}. Thus, one could argue that the improved performance of \expertconnect is merely the result of a larger learning capacity. To disprove this hypothesis we report the results for another version of the base models named {\em ``Base Model V2"}. These base models were built by increasing uniformly the number of filters and the number of units in the convolutional layers and the first fully connected layer of {\em ``Base Model V1"} in order to match exactly the total number of parameters in the \expertconnect models. \expertconnect with the same number of parameters and overall depth achieves much better accuracy than {\em ``Base Model V2"}. In subsection~\ref{sec:regularization} we will show that this is due to a regularization effect induced by our architecture. In Appendix~\ref{appendixA} we also report results obtained by shrinking the numbers of parameters in \expertconnect to match those in the original {\em ``Base Models V1''}. Even in this scenario, \expertconnect consistently outperforms the base models.

%
%

\begin{figure*}[ht]
\begin{center}
\centerline{\includegraphics[width=0.7\linewidth]{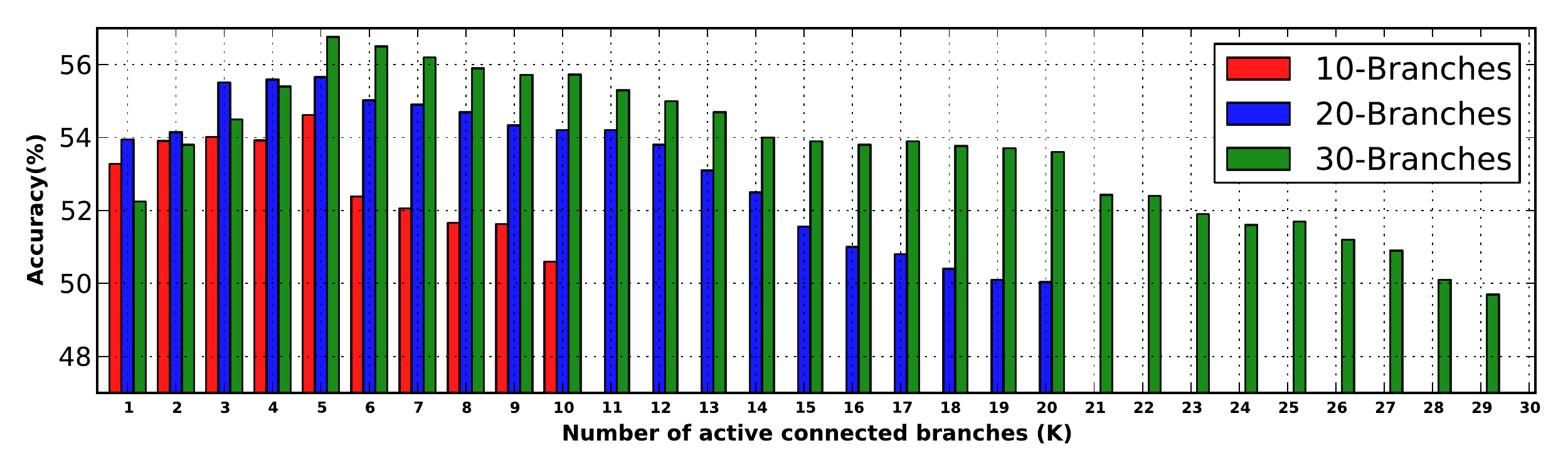}}
\caption{Accuracy of  \expertconnect using different number of branches ($M=\{$10, 20, 30$\}$) and number of active connections ($K$) on CIFAR-100. Models were built from {\em AlexNet-Quick}.}
\label{cifar100_alexquick_topmax}
\end{center}
\vspace{-0.9cm}

\end{figure*}


\subsubsection{Varying the number of branches ($M$) and the number of active connections ($K$)} 

In this subsection we study the effect of varying the number of branches ($M$) in addition to the number of active connections ($K$). Due to lack of space, here we report results only using the {\em AlexNet-Quick} base architecture but we found the overall trend on this base model to generalize also to other architectures. Figure~\ref{cifar100_alexquick_topmax} shows the accuracy achieved by \expertconnect for $M=10$, $M=20$ and $M=30$ branches. For each of these 3 architectures, we retrained our model using a varying number of active connections ($K$) ranging from 1 to $M$. It can be seen that best performance is achieved with the model having $M=30$ branches and that the model with $M=20$ branches does better than that using $M=10$ branches. This makes intuitive sense, as increasing the number of branches allows our model to further diversify the features used by the different classes. However, we can notice that in all three cases (10, 20, 30 branches), the peak accuracy is obtained at or around $K= 5$.  Moving away from this peak point, the accuracy drops nearly monotonically. When the number of active connections $K$ is near or equal to the maximum value ($M$), the models perform poorly compared to the case where $K=5$ branches are connected.  The case where only one branch is connected ($K= 1$) also gives inferior results.

\subsubsection{\expertconnectbold acts as a regularizer}
\label{sec:regularization}

We  observed in Table~\ref{table:results_cifar100} \expertconnect nets achieve higher accuracy than ``Base Models V2'' despite having the same depths and numbers of parameters. We hypothesize that this happens because of a regularization effect induced by the branched structure of our network. While a single-column CNN performs feature extraction ``horizontally'' through the layers, \expertconnect spreads some of the feature computation ``vertically'' through the parallel branches. This implies that although \expertconnect has the same number of parameters as ``Base Model V2,'' the number of feature maps extracted through each horizontal path from the input image to the end of a branch gate is smaller. This suggests that the branch features should be less prone to overfit. We confirm this hypothesis through two experiments.

In Figure~\ref{fig:cifar100_trainlosstestloss} we plot the training loss versus the test loss of four different models while varying the number of training iterations. The four models were trained on CIFAR-100 and built from {\em AlexNet-Quick} (Appendix~\ref{appendixA} includes the plot for {\em AlexNet-Full}). The four models are: 1) ``Base Model V1", 2) ``Base Model V2", 3) our \expertconnect using $K=1$ active connections out of $M=10$ branches, 4) a new model ({\em ``Random Connect"}) with $M=10$ branches where each class was permanently connected to only $K=1$ branch chosen at random from the given 10 branches. For the same training loss value, \expertconnect systematically yields lower test loss value than the base model and ``Random Connect." Also, upon convergence \expertconnect achieves lower test loss but higher training loss compare to the other two models. These results support a {\em regularization} explanation for the effect of \expertconnect. Furthermore, the overall poor test accuracy achieved by ``Random Connect" emphasizes the importance of learning the active connections rather than hardcoding them a priori.

In a second experiment, we study the effect of increasing network depth on test accuracy. In the case of single-column CNNs, if the net has already enough learning capacity, further increasing its depth makes it prone to poor local minima~\cite{erhanjmlr2010}. But if \expertconnect acts as a regularizer, it should be more resilient to increasing depths. This is confirmed by the box plots in Figure~\ref{fig:cifar100_alexquick_boxplot}, which were obtained by training both the base model ({\em AlexNet-Quick} ``Base Model V1") and the corresponding \expertconnect model (using $M=10$ and $K=5$) starting from $7$ different initializations. The figure shows the resulting distribution of test accuracy values for each model when the depth is increased up to 4 layers (top and bottom quartiles in box, maximum and minimum values above and below the box). For the base model, the accuracy goes down sharply as the depth is increased. It should also be noted that we were unable to train a base model with 4 additional layers. On the other hand, we can notice that the \expertconnect models are much more robust to the increase in depth. We were also able to train with good performance a \expertconnect model with 4 extra layers. Furthermore the variance in test accuracy is consistently lower for the case of \expertconnect for all depths. 

\begin{figure}[ht!]
\begin{center}
%
   \includegraphics[width=0.9\linewidth]{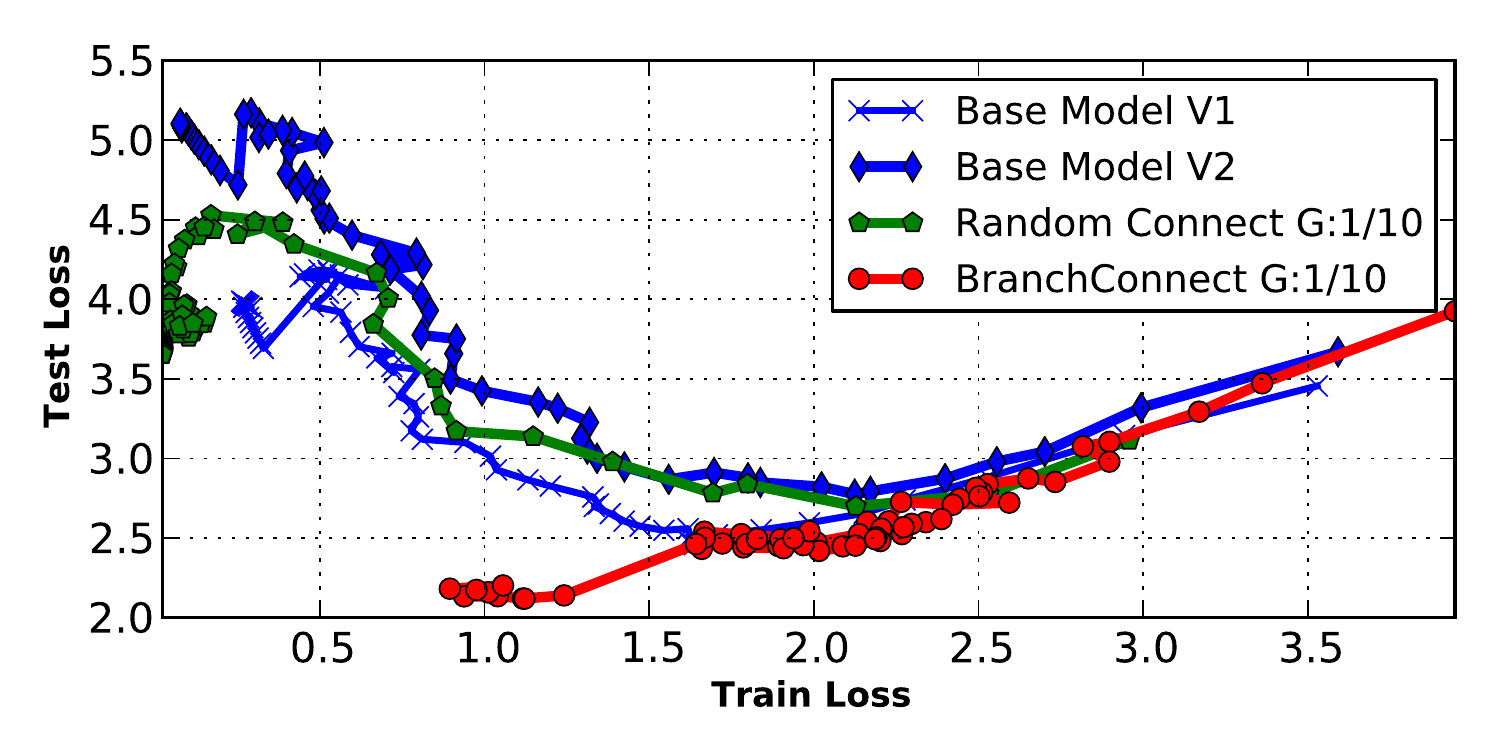}
   \label{fig:cifar100_alexquick_trainlosstestloss} 
%

\caption[]{Evolution of train loss vs test loss during training on CIFAR-100. The training trajectory is from right to left. \expertconnect yields lower test loss for the same train loss compared to the base models and a net with randomly-chosen active connections.}

\label{fig:cifar100_trainlosstestloss} 

\end{center}
\vspace{-0.6cm}

\end{figure}



\begin{figure}[ht]
\begin{center}
\centerline{\includegraphics[width=0.9\linewidth]{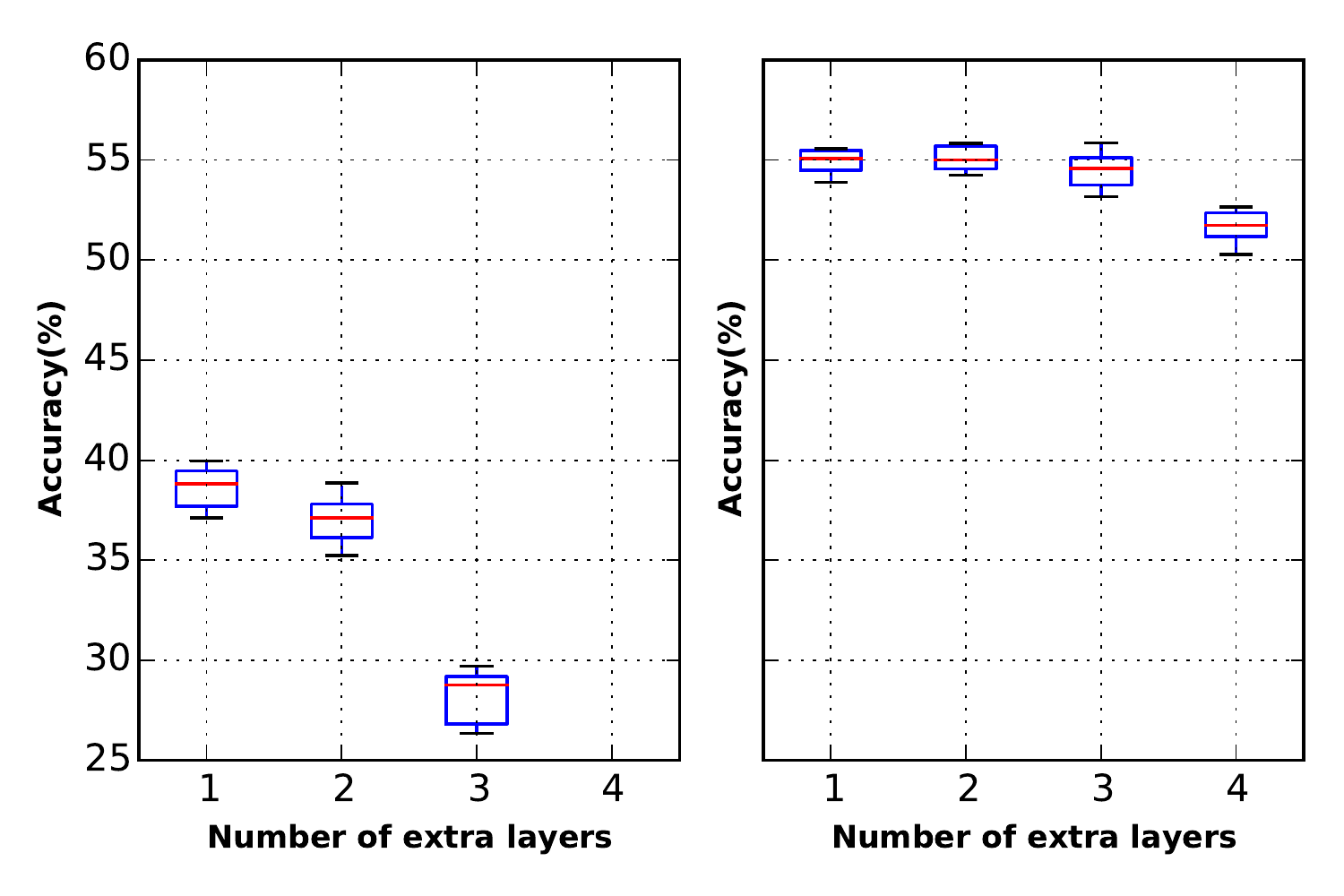}}
\caption{The effect of increasing the network depth on test accuracy (using CIFAR-100).  Boxes show the distribution of test accuracy values of the same model trained using different initialization seeds. Left: base model ({\em AlexNet-Quick}). Right: \expertconnect. Our model is less sensitive to variations in the random initialization and yields stable performance even for increased depth. }
\label{fig:cifar100_alexquick_boxplot}
\end{center}
\vspace{-0.6cm}

\end{figure}


\subsection{CIFAR-10}
In this subsection we test our approach on CIFAR-10, which includes 10 classes. This dataset is divided into 50,000 images for training, and 10,000 images for testing. Table~\ref{table:results_cifar10} shows the classification accuracy of our \expertconnect models based on three different base architectures.

\begin{table}[t]
\caption{Classification accuracy (\%) on CIFAR-10 dataset.  \textsc{G:}$K$/$M$ denotes $K$ active connections out of a total of $M$.}
\begin{center}
\label{table:results_cifar10}
{\footnotesize 
\begin{tabular}{|c||l|c|c|}
\hline
Architecture & Method & Accuracy  \\
\hline\hline
\multirow{2}{*}{AlexNet-Quick} &
Base Model    & 76.86  \\
&\expertconnect G:3/5 & \textbf{82.84} \\
\hline
\multirow{2}{*}{AlexNet-Full} &
Base Model   & 82.78  \\
& \expertconnect G:3/5 & \textbf{85.00} \\
\hline

\multirow{2}{*}{ResNet-56~\cite{he2016deep}} &
Base Model   & 92.04  \\
& \expertconnect G:3/5 & \textbf{92.46} \\
\hline
\end{tabular}
}
\end{center}
\vspace{-0.8cm}
\end{table}


\subsection{ImageNet}
\label{subsec:ImageNetExps}
We evaluate our approach on the ImageNet 2012 large-scale classification dataset~\cite{ImageNet}, which includes images of 1000 classes. The training set contains 1.28M images. We use the validation set which consist of 50K images to evaluate the performance. In Table~\ref{table:imagenetResults}, we report the Top-1 accuracies of different models.

\begin{table}[t!]
\caption{Top-1 single crop validation accuracy (\%) on ImageNet. \textsc{G:}$K$/$M$ denotes $K$ active connections out of $M$.}
\begin{center}\
\setlength{\tabcolsep}{4pt} 
\renewcommand{\arraystretch}{1}
\label{table:imagenetResults}
{\footnotesize 
\begin{tabular}{|c||l|c|c|}
\hline
Architecture & Method & Accuracy  \\
\hline\hline
\multirow{2}{*}{AlexNet~\cite{ImageNet}} &
Base Model    & 58.71 \\
& \nofe  \cite{AhmedEtAl:ECCV16} & 61.29\\
& \expertconnect G:5/10 & \textbf{63.49}\\

\hline
\multirow{2}{*}{ResNet50~\cite{he2016deep}} &
Base Model~\cite{he2016deep}    &  76.15  \\
&\expertconnect G:5/10  & \textbf{77.39} \\
&\expertconnect G:8/15   & \textbf{77.68} \\
\hline

\hline
\multirow{2}{*}{ResNet101~\cite{he2016deep}} &
Base Model~\cite{he2016deep}    &  77.37  \\
&\expertconnect G:5/10  & \textbf{78.19} \\
\hline

\end{tabular}
}
\end{center}
\vspace{-0.9cm}
\end{table}

%
%
%
%
%
%
%
%
%
%
%
%
%
%
%
%

\subsection{Synth dataset}
Finally, we evaluate our approach on a text recognition task using the Synth dataset~\cite{Jaderberg14d}. The dataset contains a total of 9M images of size 32x100. Each image contains a word drawn from a 90K dictionary. The dataset is divided into 900K images for testing, 900K images for validation, and the remaining of the images are used for training. The recognition task is to classify each of the 900K testing images into one of the 90K words (i.e., $C=90K$). The very large number of classes renders this dataset an interesting benchmark to test our \expertconnect approach. Our branched models are built from the base architecture ``DICT+2-90K" used by Jaderberg et al.~\cite{Jaderberg14c}. This base architecture has 5 convolutional layers and 3 fully-connected layers. 
Due to the large number of the classes, the training of these models was performed by adding the classes incrementally as described in~\cite{Jaderberg14c}. In Table~\ref{table:results_synth} we show the test accuracy of the base architecture and the \expertconnect models. Additionally, we show the results of the models learned from the Synth dataset when tested on other smaller datasets: IC03~\cite{IC03R16}, SVT~\cite{SVTR26}, and IC13~\cite{IC13R13}.  


\begin{table}[ht]
\caption{Word recognition accuracy (\%) for models trained on Synth, using {DICT+2-90k}~\cite{Jaderberg14c} as base model.}
\begin{center}
\setlength{\tabcolsep}{3pt} 
\renewcommand{\arraystretch}{1}
\label{table:results_synth}
{\footnotesize 
\begin{tabular}{|l|c|c|c|c|}
\hline

\multicolumn{1}{|c|}{Model} & \multicolumn{4}{|c|}{Test Dataset}  \\ 
 \cline{2-5}

 &{\scriptsize Synth~\cite{Jaderberg14d}}&{\scriptsize IC03~\cite{IC03R16}}&{\scriptsize SVT~\cite{SVTR26}}&{\scriptsize IC13~\cite{IC13R13}}\\
\hline\hline
{Base Model}~\cite{Jaderberg14d} & 95.2 & 93.1 & 80.7 & 90.8 \\
{\expertconnect G:7/10} & 95.6 & 93.7 & 83.4 & 92.1 \\
\hline
\end{tabular}
}
\end{center}

\vspace{-0.9cm}
\end{table}



\section{Conclusions}

In this paper we presented \expertconnect{---}a multi-branch, gated architecture that enables the learning of separate features for each class in large-scale classification problems. The training of our approach is end-to-end and it is posed as a single optimization that simultaneously learns the network weights and the branch connections for each class. We demonstrated the benefits of our method by adapting several popular CNNs into the form of \expertconnect. We also provided empirical analysis suggesting that \expertconnect induces a beneficial form of regularization, reducing overfitting and improving generalization.

Future work will focus on more sophisticated combination schemes. The learned gates in our model can be viewed as performing a rudimentary form of architecture learning, limited to the last layer. We plan to study the applicability of this mechanism for more general forms of model learning. 

{\small
\bibliographystyle{ieee}
\bibliography{egbib}
}

\clearpage

\appendix

\section{Appendix}\label{appendixA}

This supplementary material is organized as follows: in subsection~\ref{exps_section}, we show additional experiments on CIFAR-100; in subsection~\ref{reg_section} we show additional plots that further support the interpretation of \expertconnect as a regularizer; in subsection~\ref{spec_section} we provide the specifications of the networks used in the paper including the details of the training process and data preprocessing.

\subsection{Additional Experiments on CIFAR-100}
\label{exps_section}
In Table~\ref{table:results_cifar100} we report additional experiments on the CIFAR-100 dataset. In the main paper we presented results of \expertconnect networks having the same number of parameters as {\em ``Base Models V2''} models, which are larger capacity versions of {\em ``Base Models V1''}. Here we report results of additional \expertconnect models ({\em ``\expertconnect V1"}) that are obtained by shrinking the numbers of parameters to match those in the original {\em ``Base Models V1''}. Once again, we see that \expertconnect networks consistently outperform the base models of the same capacity, both in the case of {\em ``Base Models V1''} as well as {\em ``Base Models V2''}.

\subsection{Regularization Effect Induced by BranchConnect}
\label{reg_section}
In the main paper we provided a figure that shows the training loss versus the test loss of different models that were trained on CIFAR-100 and built from {\em AlexNet-Quick} architecture (Figure 3 in the paper). Here we show the training loss versus the test loss of different models that were built from other architectures.

\begin{figure*}[t]
\begin{center}
	\begin{subfigure}[]{0.7\linewidth}
      \includegraphics[width=\linewidth]{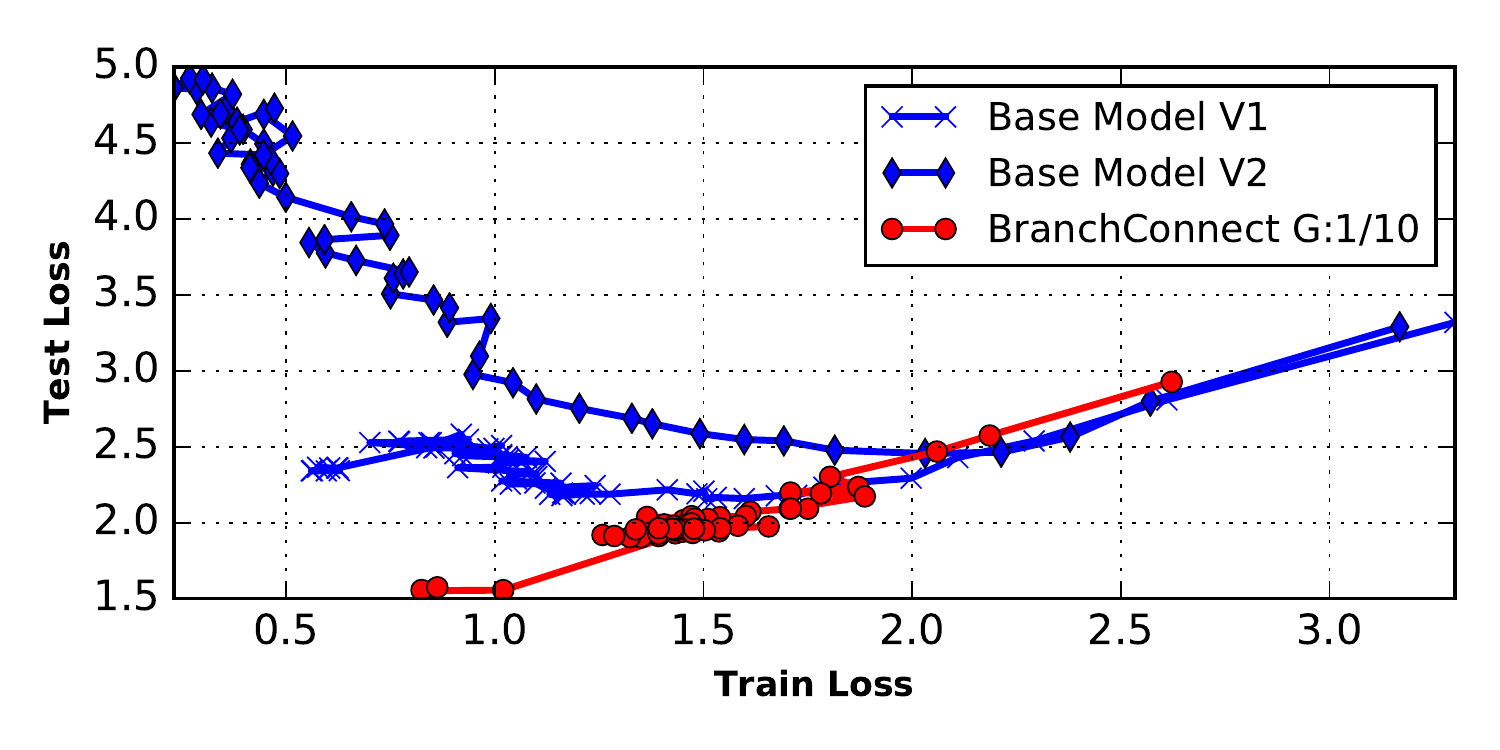}
   \caption{{\bf AlexNet-Full}}
   \label{fig:cifar100_trainlosstestloss_a} 
\end{subfigure}
\begin{subfigure}[]{0.7\linewidth}
   \includegraphics[width=\linewidth]{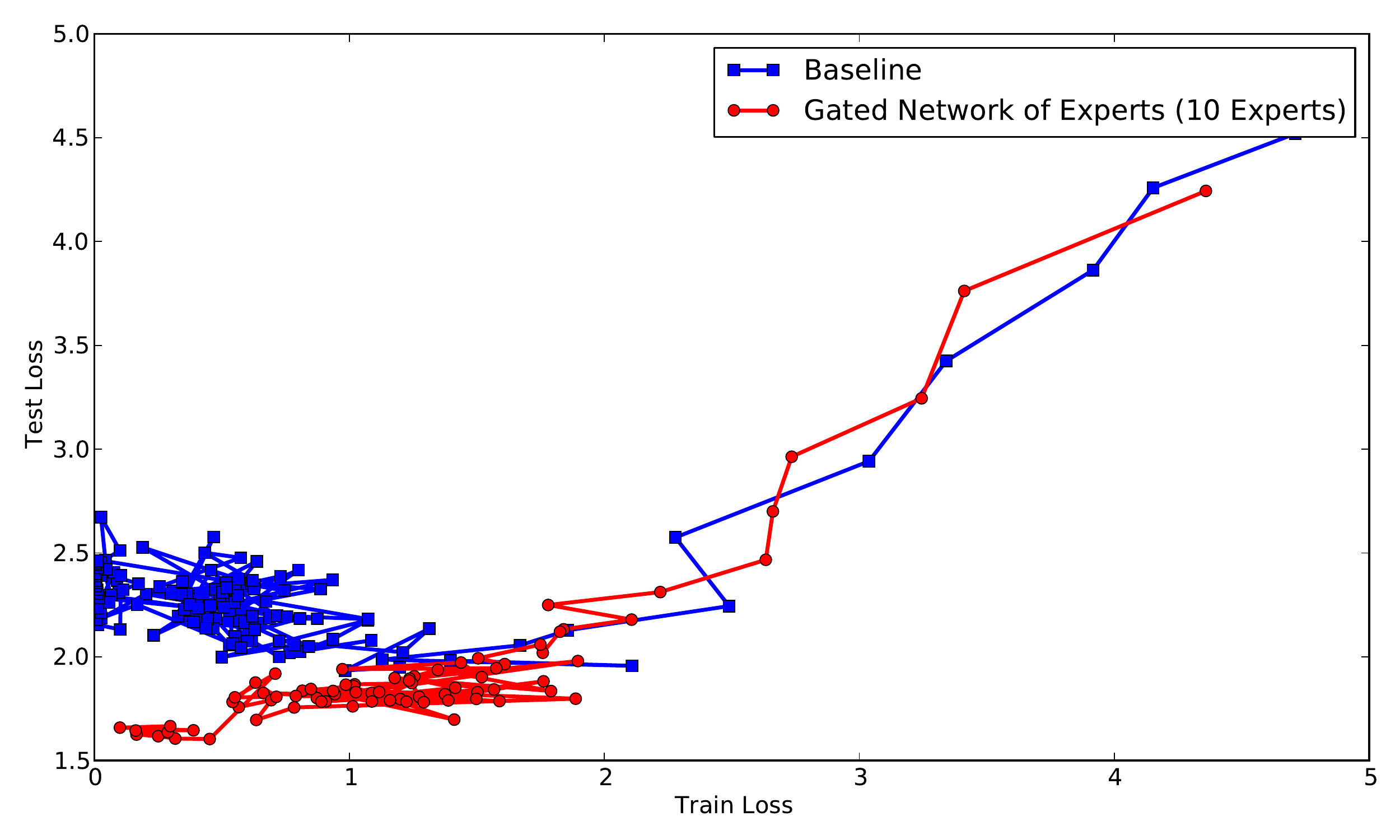}
   \caption{{\bf Network In Network (NIN)}}
   \label{fig:cifar100_trainlosstestloss_b} 
\end{subfigure}


\caption[]{Evolution of train loss vs test loss during training on CIFAR-100. The models were built from different architectures: a) AlexNet-Full architecture, b) Network In Network (NIN) architecture. The training trajectory is from right to left. \expertconnect yields lower test loss for the same train loss compared to the base models and a net with randomly-chosen active connections.}

\label{fig:cifar100_trainlosstestloss} 

\end{center}

\end{figure*}

 In Figure~\ref{fig:cifar100_trainlosstestloss_a} we show the training loss versus the test loss for three different models that were trained on CIFAR-100 and built from the {\em AlexNet-Full} architecture: 1) ``Base Model V1", 2) ``Base Model V2", 3) our \expertconnect using $K=1$ active connection out of $M=10$ branches. In Figure~\ref{fig:cifar100_trainlosstestloss_b} we show the training loss versus the test loss for two models that were trained on CIFAR-100 and built from the {\em NIN} architecture:  1) the \expertconnect model using $K=1$ active connection out of $M=10$ branches, 2) ``Base Model V2" which has the same number of parameters as the \expertconnect network,
In all figures, we notice that for the same training loss value, our \expertconnect yields lower test loss value than the base models. Additionally, we notice that upon convergence \expertconnect achieves lower test loss but higher training loss compared to the other models. These results support a {\em regularization} explanation for the effect of \expertconnect.

\subsection{Network Specifications}
\label{spec_section}
In this subsection, we provide the network specification and the learning policy of the models used in our experiments on CIFAR100, CIFAR10, ImageNet and Synth datasets. Tables~\ref{table:alex_quick_config},~\ref{table:alex_full_config},
~\ref{table:nin_config},~\ref{table:resnet1_config}, and~\ref{table:resnet2_config} list the specifications of the original base models (``Base Model V1" in the paper) and the \expertconnect models built from the five architectures that were used to train CIFAR100 dataset: AlexNet-Quick, AlexNet-Full, NIN, Resnet56, and Resnet56-4X  (classification accuracy of different models are given in Table 1 of the paper). 

For training on CIFAR10, we used the same architectures presented in Tables~\ref{table:alex_quick_config},~\ref{table:alex_full_config}, and~\ref{table:resnet2_config} except that the number of output units in the last fully-connected layer was set to 10 instead of 100 (classification accuracy of different models are given in Table 2 of the paper). 

For training on ImageNet, we used the network specification and the learning policy as listed in Tables~\ref{table:imagenet_alexnet_config},~\ref{table:imagenet_resnet50_config}, and~\ref{table:imagenet_resnet101_config}.

In Table~\ref{table:synth_config} we show the the specifications of the base model and the corresponding \expertconnect model used on the Synth dataset.

Each table lists the specification in two columns starting from top (input) to down (output). The first column shows the architecture and the training policy of the base model, while the second column shows the architecture and the training policy of the corresponding \expertconnect model. For all of the \expertconnect models, we show the layers included in the stem of \expertconnect, and the layers included in the branches. We used rectified linear units (ReLU) in all architectures. In the second part of each table, we show the input data preprocessing done before training each model. The third part shows the learning policy for each model. 

\subsection*{Notation used to describe layers:}
\begin{itemize}
\item \textbf{Convolutional layer:}\\ {\footnotesize CONV: $<$filter width$>\times<$filter height$>,<$number of filters$>$}  \\
 Example:  { CONV: 5$\times$5,32}  means a convolutional layer with 32 filters of size 5$\times$5. 

\item \textbf{Pooling layer:}\\ {\footnotesize POOL: $<$filter width$>\times<$filter height$>$,$<$type$>$,$<$stride$>$ } \\ 
 Example:  { POOL: 3$\times$3,Ave,2} denotes an average pooling layer of filter size 3$\times$3 and stride 2. Two types are used: ``Max" means maximum pooling, and ``Ave" means Average pooling. 

\item \textbf{LRN:} Local Response Normalization layer. 

\item \textbf{Fully Connected layer:}\\
{\footnotesize FC: $<$number of output units$>$}\\
 Example:  { FC:$100$} indicates a fully connected layer with 100 output units.

\item \textbf{Fully Connected layer with Gates:}\\
{\footnotesize FC\_Gates: $<$number of output units$>$}\\
 As described in the paper, this type of fully-connected layer includes gates and it is used to build the \expertconnect models. For example, { FC\_Gates:$100$} indicates a fully connected layer with 100 output units and also 100 gates, where each gate is connected to each output unit. \\

\item \textbf{Convolution Layer with Gates:}\\
{\footnotesize {CONV\_Gates: $<$filter width$>\times<$filter height$>,<$number of filters$>$}}\\
Similar to fully-connected layer with gates, this is layer is connected to gates and it is used as an output layer for the \expertconnect models that were built using NIN architecture.   

\item \textbf{Residual block:} \\
Residual blocks are the building units of the Residual Networks~\cite{he2016deep}. In Tables~\ref{table:resnet1_config} and~\ref{table:resnet2_config} we use big braces to refer to a residual block. For instance, the following: $ 
\left \{
  {\footnotesize 
    \begin{tabular}{l}
 {CONV: 3$\times$3,64} \\
 {CONV: 3$\times$3,64}
  \end{tabular}
    }
\right \}\times9
$    , denotes a group of $9$ residual blocks where each block consists of two layers {CONV: 3$\times$3,64}, with batch normalization and scaling layers as proposed in~\cite{he2016deep}.

\end{itemize}


\begin{table}[t!]
\caption{Classification accuracy (\%) (single crop) on CIFAR-100 for 3 different base architectures. {\em ``Base Models V1''}  represent the original model architectures from prior work~\cite{NIN, AhmedEtAl:ECCV16}, and {\em ``\expertconnect V1''} are the equivalent \expertconnect models obtained by shrinking  the numbers of parameters in \expertconnect to match those in these original base models. Conversely, {\em ``\expertconnect V2''} represent higher-capacity \expertconnect models whose branches have the same specifications as the last layer(s) of {\em ``Base Models V1''}. In order to compare  {\em ``\expertconnect V2''} with base models of the same capacity, we build  {\em ``Base Models V2''} obtained from {\em ``Base Models V1''} by increasing uniformly the number of filters and the number of units. These results show that \expertconnect networks consistently outperform base models of the same capacity.}

\small{
\begin{center}\
\setlength{\tabcolsep}{4pt} 
\renewcommand{\arraystretch}{1}
\label{table:results_cifar100}

\begin{tabular}{|p{0.2cm}||l|C{0.8cm}|C{1.2cm}|C{1.35cm}|}

\hline
 & Method & depth & \#params & Accuracy  \\
\hline\hline

\multirow{5}{*}{\rotatebox[origin=c]{90}{\bf \footnotesize  AlexNet-Quick}} &
Base Model V1 & 5 & 0.15M  & 44.3\\

& \expertconnect G:1/10 V1 & 5 & 0.15M & \textbf{49.14}\\
& \expertconnect G:5/10 V1 & 5 & 0.15M & \textbf{52.81}\\

\cline{2-5}

& Base Model V2 & 5 & 1.20M & 40.26\\

& \expertconnect G:1/10 V2 & 5 & 1.20M & \textbf{53.28}\\
& \expertconnect G:5/10 V2 & 5 & 1.20M & \textbf{54.62}\\

\hline\hline

\multirow{5}{*}{\rotatebox[origin=c]{90}{\bf \footnotesize Alexnet-Full}} &

Base Model V1 & 4& 0.18M & 54.04  \\
& \expertconnect G:1/10 V1 & 4 & 0.18M & \textbf{55.61} \\
& \expertconnect G:6/10 V1 & 4 & 0.18M & \textbf{56.73} \\

\cline{2-5}

& Base Model V2 & 4& 0.64M & 50.42   \\
& \expertconnect G:1/10 V2 & 4 & 0.64M & \textbf{57.34} \\
& \expertconnect G:6/10 V2 & 4 & 0.64M & \textbf{60.27} \\

\hline\hline

\multirow{5}{*}{\rotatebox[origin=c]{90}{\bf \footnotesize NIN~\cite{NIN}}} &
Base Model V1 & 9 & 1.38M & 64.73  \\

&\expertconnect G:1/10 V1 & 9 & 1.38M  & \textbf{65.46} \\
&\expertconnect G:5/10 V1 & 9 & 1.38M  & \textbf{65.99} \\

\cline{2-5}

&Base Model V2 & 9 & 1.61M & 65.24  \\

&\expertconnect G:1/10 V2 & 9 & 1.61M  & \textbf{66.10} \\
&\expertconnect G:5/10 V2 & 9 & 1.61M  & \textbf{66.45} \\

\hline

\end{tabular}
\end{center}
}
\end{table}


\begin{table*}[t]
\begin{center}
\caption{\bf CIFAR100 AlexNet-Quick}
\label{table:alex_quick_config}

\begin{tabular}{ |C{4.5cm}|| p{0.2cm} |C{3.8cm}|  }
\hline
\bf{Base Model} &  \multicolumn{2}{|c|}{\expertconnectbold} \\ 
\hline
\hline

 {CONV: 5$\times$5,32 } &  \multirow{4}{*}{\rotatebox[origin=c]{90}{\bf  Stem}} & CONV: 5$\times$5,32  \\ 
 {POOL: 3$\times$3,Max,2}  & & POOL: 3$\times$3,Max,2   \\

 \cline{1-1}\cline{3-3} 
{CONV: 5$\times$5,32}  & & CONV: 5$\times$5,32  \\ 
{POOL: 3$\times$3,Ave,2}  & & POOL: 3$\times$3,Ave,2   \\
\hline

{CONV: 5$\times$5,64}  &  \multirow{3}{*}{\rotatebox[origin=c]{90}{\bf Branch}}  & {CONV:  5$\times$5,64}  \\ 
{POOL: 3$\times$3,Ave,2}  & & POOL: 3$\times$3,Ave,2   \\

\cline{1-1}\cline{3-3} 

{FC: 64 } 	&           &   {FC: 64 }   \\ 
\hline
{FC: 100 } 	&           \multicolumn{2}{c|} {FC\_Gates: 100 }   \\

\hline
\hline
\multicolumn{3}{|c|}{\bf Data preprocessing} \\
\hline
\parbox[t]{4cm}{Input: $32\times32$\\Image mean subtraction} 
& 
\multicolumn{2}{|l|}{\parbox{4cm}{Input: $32\times32$\\Image mean subtraction}}
\\
\hline
\multicolumn{3}{|c|}{\bf Learning Policy} \\
\hline

\parbox{6cm}{
{\footnotesize 
$\bullet$ Learning rate: 0.001, 0.0001, 0.00001\\ 
$\bullet$ Momentum: 0.9\\
$\bullet$ Weight decay: 0.004\\ 
$\bullet$ Weight initialization: Random\\ 
} }
& 
\multicolumn{2}{l|}{\parbox{5.5cm}{
{\footnotesize 
$\bullet$ Learning rate: 0.001, 0.0001, 0.00001\\ 
$\bullet$ Gate Learning rate: 10$\times$Learning rate \\
$\bullet$ Momentum: 0.9\\
$\bullet$ Weight decay: 0.004\\
$\bullet$ Weight initialization: Random
}}}
 
\\
 
\hline 
\end{tabular}

\end{center}
\end{table*}


\begin{table*}
\begin{center}
\caption{\bf CIFAR100  AlexNet-Full}
\label{table:alex_full_config}

\begin{tabular}{ |C{4.5cm}|| p{0.2cm} |C{3.8cm}|  }
\hline
\bf{Base Model} &  \multicolumn{2}{|c|}{\expertconnectbold} \\ 
\hline
\hline

{CONV: 5$\times$5,32 } &  \multirow{4}{*}{\rotatebox[origin=c]{90}{\bf  Stem}} & CONV: 5$\times$5,32  \\ 
{POOL: 3$\times$3,Max,2}  & & POOL: 3$\times$3,Max,2   \\
{LRN}  &  & LRN    \\

 \cline{1-1}\cline{3-3} 
{CONV: 5$\times$5,32}  & & CONV: 5$\times$5,32  \\ 
{POOL: 3$\times$3,Ave,2}  & & POOL: 3$\times$3,Ave,2   \\
{LRN}  &  & LRN    \\
\hline

{CONV: 5$\times$5,64}  &  \multirow{2}{*}{\rotatebox[origin=c]{90}{\bf Branch}}  & {CONV: 5$\times$5,64}  \\ 
{POOL: 3$\times$3,Ave,2}  & & POOL: 3$\times$3,Ave,2   \\
& & \\
\hline

{FC: 100 } 	&           \multicolumn{2}{c|} {FC\_Gates: 100 }   \\

\hline
\hline
\multicolumn{3}{|c|}{\bf Data preprocessing} \\
\hline
\parbox[t]{4cm}{Input: $32\times32$\\Image mean subtraction} 
& 
\multicolumn{2}{|l|}{\parbox{4cm}{Input: $32\times32$\\Image mean subtraction}}
\\
\hline
\multicolumn{3}{|c|}{\bf Learning Policy} \\
\hline

\parbox{6cm}{
{\footnotesize 
$\bullet$ Learning rate: 0.001, 0.0001, 0.00001\\ 
$\bullet$ Momentum: 0.9\\
$\bullet$ Weight decay: 0.004\\ 
$\bullet$ Weight initialization: Random\\ 
} }
& 
\multicolumn{2}{l|}{\parbox{5.5cm}{
{\footnotesize 
$\bullet$ Learning rate: 0.001, 0.0001, 0.00001\\ 
$\bullet$ Gate Learning rate: 10$\times$Learning rate \\
$\bullet$ Momentum: 0.9\\
$\bullet$ Weight decay: 0.004\\
$\bullet$ Weight initialization: Random
}}}
 
\\
 
\hline 
\end{tabular}

\end{center}
\end{table*}


\begin{table*}
\begin{center}
\caption{\bf CIFAR100 - Network In Network (NIN)}
\label{table:nin_config}

\begin{tabular}{ |C{4.5cm}|| p{0.2cm} |C{3.8cm}|  }
\hline
\bf{Base Model} &  \multicolumn{2}{|c|}{\expertconnectbold} \\ 
\hline
\hline

 {CONV: 3$\times$3,192 } &  \multirow{4}{*}{\rotatebox[origin=c]{90}{\bf  Stem}} & CONV: 3$\times$3,192  \\ 
{CONV: 3$\times$3,184}  & & {CONV: 3$\times$3,184}   \\
{CONV: 1$\times$1,158}  & & {CONV: 1$\times$1,158}   \\
{POOL: 3$\times$3,Max,2}  & & POOL: 3$\times$3,Max,2   \\

 \cline{1-1}\cline{3-3} 
{CONV: 3$\times$3,192}  & & {CONV: 3$\times$3,192}   \\
{CONV: 3$\times$3,192}  & & {CONV: 3$\times$3,192}   \\
{CONV: 1$\times$1,192}  & & {CONV: 1$\times$1,192}   \\
{POOL: 3$\times$3,Max,2}  & & POOL: 3$\times$3,Max,2   \\
{CONV: 3$\times$3,192}  & & {CONV: 3$\times$3,192}   \\
  & & POOL: 3$\times$3,Ave,2   \\

 \hline
 
{CONV: 1$\times$,192}  &  \multirow{3}{*}{\rotatebox[origin=c]{90}{\bf Branch}}  & {CONV: 1$\times$,192} \\ 
& & \\
& & \\


\hline
{CONV: 1$\times$1,100} 	&       \multicolumn{2}{|c|}{CONV\_Gates: 1$\times$1,100}  \\ 
{POOL: 1$\times$1,Ave,11}  &   \multicolumn{2}{|c|}{POOL: 1$\times$1,Ave,11}   \\

\hline
\hline
\multicolumn{3}{|c|}{\bf Data preprocessing} \\
\hline
\parbox[t]{4cm}{Input: $32\times32$\\Image mean subtraction} 
& 
\multicolumn{2}{|l|}{\parbox{4cm}{Input: $32\times32$\\Image mean subtraction}}
\\
\hline
\multicolumn{3}{|c|}{\bf Learning Policy} \\
\hline

\parbox{6cm}{
{\footnotesize 
$\bullet$ Learning rate: 0.025, 0.0125, 0.0001\\ 
$\bullet$ Momentum: 0.9\\
$\bullet$ Weight decay: 0.0005\\ 
$\bullet$ Weight initialization: Random\\ \\
} }
& 
\multicolumn{2}{|l|}{\parbox{5.5cm}{
{\footnotesize 
$\bullet$ Learning rate: 0.025, 0.0125, 0.0001\\ 
$\bullet$ Gate Learning rate: 10$\times$Learning rate \\
$\bullet$ Momentum: 0.9\\
$\bullet$ Weight decay: 0.0005\\
$\bullet$ Weight initialization: Random
}}}
 
\\
 
\hline 
\end{tabular}

\end{center}
\end{table*}


\begin{table*}
\begin{center}
\caption{\bf CIFAR100 ResNet56-4X} 
\label{table:resnet1_config}

\begin{tabular}{ |C{5cm}|| p{0.2cm} |C{5cm}|  }
\hline
\bf{Base Model} &  \multicolumn{2}{c|}{\expertconnectbold} \\ 
\hline
\hline

{CONV: 3$\times$3,64} &  \multirow{4}{*}{\rotatebox[origin=c]{90}{\bf  Stem}} & CONV: 3$\times$3,64 \\ 

\cline{1-1}\cline{3-3} 

{ \Bigg\{ \parbox[]{2.7cm}{CONV: 3$\times$3,64 \\ CONV: 3$\times$3,64}  \Bigg\}   $\times9$  } 
& &
{ \Bigg\{ \parbox[]{2.7cm}{CONV: 3$\times$3,64 \\ CONV: 3$\times$3,64}  \Bigg\}   $\times9$  }   \\ 

\cline{1-1}\cline{3-3}

{ \Bigg\{ \parbox[]{2.7cm}{CONV: 3$\times$3,128 \\ CONV: 3$\times$3,128}  \Bigg\}   $\times9$  } 
& &
{ \Bigg\{ \parbox[]{2.7cm}{CONV: 3$\times$3,128 \\ CONV: 3$\times$3,128}  \Bigg\}   $\times9$  }   \\ 

\cline{1-1}\cline{3-3}

{ \Bigg\{ \parbox[]{2.7cm}{CONV: 3$\times$3,256 \\ CONV: 3$\times$3,256}  \Bigg\}   $\times9$  } 
& &
{ \Bigg\{ \parbox[]{2.7cm}{CONV: 3$\times$3,256 \\ CONV: 3$\times$3,256}  \Bigg\}   $\times8$  }   \\

\hline

{POOL: 7$\times$7,Ave,1} &   \multirow{2}{*}{\rotatebox[origin=c]{90}{\bf Branch}}  & 
{ \Bigg\{ \parbox[]{2.7cm}{CONV: 3$\times$3,256 \\ CONV: 3$\times$3,256}  \Bigg\}   $\times1$ }  \\ 
  &  & {POOL: 7$\times$7,Ave,1}    \\

\hline
{FC: 100 } 	&          \multicolumn{2}{c|}  {FC\_Gates: 100 }   \\

\hline
\hline
\multicolumn{3}{|c|}{\bf Data preprocessing} \\
\hline
\parbox[t]{5cm}{Input: $28\times28$\\Image mean subtraction\\Image mirroring} 
& 
\multicolumn{2}{l|}{\parbox{4cm}{Input: $28\times28$\\Image mean subtraction\\Image mirroring}}
\\
\hline
\multicolumn{3}{|c|}{\bf Learning Policy} \\
\hline

\parbox{5.5cm}{
{\footnotesize 
$\bullet$ Learning rate: 0.1, 0.01, 0.001\\ 
$\bullet$ Momentum: 0.9\\
$\bullet$ Weight decay: 0.0001\\ 
$\bullet$ Weight initialization: MSRA~\cite{HeEtAl:ICCV2015} \\
} }
& 
\multicolumn{2}{l|}{\parbox{5cm}{
{\footnotesize 
$\bullet$ Learning rate: 0.1, 0.01, 0.001\\ 
$\bullet$ Gate Learning rate: 10$\times$Learning rate \\
$\bullet$ Momentum: 0.9\\
$\bullet$ Weight decay: 0.0001\\ 
$\bullet$ Weight initialization: MSRA~\cite{HeEtAl:ICCV2015} 
}}}
 
\\
 
\hline 
\end{tabular}

\end{center}
\end{table*}


\begin{table*}
\begin{center}
\caption{\bf CIFAR100 ResNet56} 
\label{table:resnet2_config}

\begin{tabular}{ |C{5cm}|| p{0.2cm} |C{5cm}|  }
\hline
\bf{Base Model} &  \multicolumn{2}{c|}{\expertconnectbold} \\ 
\hline
\hline

{CONV: 3$\times$3,16} &  \multirow{4}{*}{\rotatebox[origin=c]{90}{\bf  Stem}} & CONV: 3$\times$3,16 \\ 

\cline{1-1}\cline{3-3} 

{ \Bigg\{ \parbox[]{2.7cm}{CONV: 3$\times$3,16 \\ CONV: 3$\times$3,16}  \Bigg\}   $\times9$  } 
& &
{ \Bigg\{ \parbox[]{2.7cm}{CONV: 3$\times$3,16 \\ CONV: 3$\times$3,16}  \Bigg\}   $\times9$  }   \\ 

\cline{1-1}\cline{3-3}

{ \Bigg\{ \parbox[]{2.7cm}{CONV: 3$\times$3,32 \\ CONV: 3$\times$3,32}  \Bigg\}   $\times9$  } 
& &
{ \Bigg\{ \parbox[]{2.7cm}{CONV: 3$\times$3,32 \\ CONV: 3$\times$3,32}  \Bigg\}   $\times9$  }   \\ 

\cline{1-1}\cline{3-3}

{ \Bigg\{ \parbox[]{2.7cm}{CONV: 3$\times$3,64 \\ CONV: 3$\times$3,64}  \Bigg\}   $\times9$  } 
& &
{ \Bigg\{ \parbox[]{2.7cm}{CONV: 3$\times$3,64 \\ CONV: 3$\times$3,64}  \Bigg\}   $\times8$  }   \\

\hline

{POOL: 7$\times$7,Ave,1} &   \multirow{2}{*}{\rotatebox[origin=c]{90}{\bf Branch}}  & 
{ \Bigg\{ \parbox[]{2.7cm}{CONV: 3$\times$3,64 \\ CONV: 3$\times$3,64}  \Bigg\}   $\times1$ }  \\ 
  &  & {POOL: 7$\times$7,Ave,1}    \\

\hline
{FC: 100 } 	&           \multicolumn{2}{c|}   {FC\_Gates: 100 }   \\

\hline
\hline
\multicolumn{3}{|c|}{\bf Data preprocessing} \\
\hline
\parbox[t]{5cm}{Input: $28\times28$\\Image mean subtraction\\Image mirroring} 
& 
\multicolumn{2}{l|}{\parbox{4cm}{Input: $28\times28$\\Image mean subtraction\\Image mirroring}}
\\
\hline
\multicolumn{3}{|c|}{\bf Learning Policy} \\
\hline

\parbox{5.5cm}{
{\footnotesize 
$\bullet$ Learning rate: 0.1, 0.01, 0.001\\ 
$\bullet$ Momentum: 0.9\\
$\bullet$ Weight decay: 0.0001\\ 
$\bullet$ Weight initialization: MSRA~\cite{HeEtAl:ICCV2015} \\
} }
& 
\multicolumn{2}{l|}{\parbox{5cm}{
{\footnotesize 
$\bullet$ Learning rate: 0.1, 0.01, 0.001\\ 
$\bullet$ Gate Learning rate: 10$\times$Learning rate \\
$\bullet$ Momentum: 0.9\\
$\bullet$ Weight decay: 0.0001\\ 
$\bullet$ Weight initialization: MSRA~\cite{HeEtAl:ICCV2015} 
}}}
 
\\
 
\hline 
\end{tabular}

\end{center}
\end{table*}


\begin{table*}
\begin{center}
\caption{\bf ImageNet  AlexNet}
\label{table:imagenet_alexnet_config}

\begin{tabular}{ |C{4.5cm}|| p{0.2cm} |C{3.8cm}|  }
\hline
\bf{Base Model} &  \multicolumn{2}{|c|}{\expertconnectbold} \\ 
\hline
\hline

{CONV: 11$\times$11,96 } &  \multirow{5}{*}{\rotatebox[origin=c]{90}{\bf  Stem}} & 11$\times$11,96  \\ 
{CONV: 11$\times$11,96 } &  & 11$\times$11,96  \\ 
{LRN}  &  & LRN    \\

{POOL: 3$\times$3,Max,2}  & & POOL: 3$\times$3,Max,2   \\

 \cline{1-1}\cline{3-3} 
 
 {CONV: 3$\times$3,384}  & & CONV: 3$\times$3,384  \\ 
 {CONV: 3$\times$3,384}  & & CONV: 3$\times$3,384  \\

  {CONV: 3$\times$3,256}  & &   \\ 
  {LRN}  &  &     \\

{POOL: 3$\times$3,Max,2}  & &  \\

 \hline

  &  \multirow{3}{*}{\rotatebox[origin=c]{90}{\bf Branch}}  & {CONV:  3$\times$3,256}  \\ 
 &  & LRN    \\

    && {POOL: 3$\times$3,Max,2} \\ 

  {FC: 4096}   &&  {FC: 1024}  \\ 
  {FC: 4096}   &&  {FC: 1024}  \\

\hline
{FC: 1000 } 	&           \multicolumn{2}{c|} {FC\_Gates: 1000 }   \\

\hline
\hline
\multicolumn{3}{|c|}{\bf Data preprocessing} \\
\hline
\parbox[t]{4cm}{Input: $227\times227$\\Image mean subtraction} 
& 
\multicolumn{2}{|l|}{\parbox{4cm}{Input: $227\times227$\\Image mean subtraction}}
\\
\hline
\multicolumn{3}{|c|}{\bf Learning Policy} \\
\hline

\parbox{6cm}{
{\footnotesize 
$\bullet$ Learning rate: 0.01, 0.001, 0.001\\ 
$\bullet$ Momentum: 0.9\\
$\bullet$ Weight decay: 0.0005\\ 
$\bullet$ Weight initialization: Random\\ 
} }
& 
\multicolumn{2}{l|}{\parbox{5.5cm}{
{\footnotesize 
$\bullet$ Learning rate: 0.01, 0.001, 0.001\\ 
$\bullet$ Gate Learning rate: 10$\times$Learning rate \\
$\bullet$ Momentum: 0.9\\
$\bullet$ Weight decay: 0.0005\\ 
$\bullet$ Weight initialization: Random
}}}
 
\\
 
\hline 
\end{tabular}

\end{center}
\end{table*}


\begin{table*}
\begin{center}
\caption{\bf ImageNet ResNet-50} 
\label{table:imagenet_resnet50_config}

\begin{tabular}{ |C{5cm}|| p{0.2cm} |C{5cm}|  }
\hline
\bf{Base Model} &  \multicolumn{2}{c|}{\expertconnectbold} \\ 
\hline
\hline

{CONV: 7$\times$7,64} &  \multirow{5}{*}{\rotatebox[origin=c]{90}{\bf  Stem}} & CONV: 7$\times$7,64 \\ 
{POOL: 3$\times$3,Max,2}  & & {POOL: 3$\times$3,Max,2} \\

\cline{1-1}\cline{3-3} 

{ \Bigg\{ \parbox[]{2.7cm}{CONV: 1$\times$1,64 \\ CONV: 3$\times$3,64 \\ CONV: 1$\times$1,256}  \Bigg\}   $\times3$  } 
& &
{ \Bigg\{ \parbox[]{2.7cm}{CONV: 1$\times$1,64 \\ CONV: 3$\times$3,64 \\ CONV: 1$\times$1,256}  \Bigg\}   $\times3$  } 
\\ 
\cline{1-1}\cline{3-3}

{ \Bigg\{ \parbox[]{2.7cm}{CONV: 1$\times$1,128 \\ CONV: 3$\times$3,128 \\ CONV: 1$\times$1,512}  \Bigg\}   $\times4$  } 
& &
{ \Bigg\{ \parbox[]{2.7cm}{CONV: 1$\times$1,128 \\ CONV: 3$\times$3,128 \\ CONV: 1$\times$1,512}  \Bigg\}   $\times4$  } 
\\ 
\cline{1-1}\cline{3-3}

{ \Bigg\{ \parbox[]{2.7cm}{CONV: 1$\times$1,256 \\ CONV: 3$\times$3,256 \\ CONV: 1$\times$1,1024}  \Bigg\}   $\times6$  } 
& &
{ \Bigg\{ \parbox[]{2.7cm}{CONV: 1$\times$1,256 \\ CONV: 3$\times$3,256 \\ CONV: 1$\times$1,1024}  \Bigg\}   $\times6$  } 
\\ 
\cline{1-1}\cline{3-3}

{ \Bigg\{ \parbox[]{2.7cm}{CONV: 1$\times$1,512 \\ CONV: 3$\times$3,512 \\ CONV: 1$\times$1,2048}  \Bigg\}   $\times3$  } 
& &
{ \Bigg\{ \parbox[]{2.7cm}{CONV: 1$\times$1,512 \\ CONV: 3$\times$3,512 \\ CONV: 1$\times$1,2048}  \Bigg\}   $\times2$  } 
\\ 
\cline{1-1}\cline{3-3}

\hline

{POOL: 7$\times$7,Ave,1} &   \multirow{2}{*}{\rotatebox[origin=c]{90}{\bf Branch}}  & 
{ \Bigg\{ \parbox[]{2.7cm}{CONV: 1$\times$1,512 \\ CONV: 3$\times$3,512 \\ CONV: 1$\times$1,1024}  \Bigg\}   $\times1$ }  \\ 
  &  & {POOL: 7$\times$7,Ave,1}    \\

\hline
{FC: 1000 } 	&           \multicolumn{2}{c|}   {FC\_Gates: 1000 }   \\

\hline
\hline
\multicolumn{3}{|c|}{\bf Data preprocessing} \\
\hline
\parbox[t]{5cm}{Input: $224\times224$\\Image mean subtraction\\Image mirroring} 
& 
\multicolumn{2}{l|}{\parbox{4cm}{Input: $224\times224$\\Image mean subtraction\\Image mirroring}}
\\
\hline
\multicolumn{3}{|c|}{\bf Learning Policy} \\
\hline

\parbox{5.5cm}{
{\footnotesize 
$\bullet$ Learning rate: 0.1, 0.01, 0.001\\ 
$\bullet$ Momentum: 0.9\\
$\bullet$ Weight decay: 0.0001\\ 
$\bullet$ Weight initialization: MSRA~\cite{HeEtAl:ICCV2015} \\
} }
& 
\multicolumn{2}{l|}{\parbox{5cm}{
{\footnotesize 
$\bullet$ Learning rate: 0.1, 0.01, 0.001\\ 
$\bullet$ Gate Learning rate: 10$\times$Learning rate \\
$\bullet$ Momentum: 0.9\\
$\bullet$ Weight decay: 0.0001\\ 
$\bullet$ Weight initialization: MSRA~\cite{HeEtAl:ICCV2015} 
}}}
 
\\
 
\hline 
\end{tabular}

\end{center}
\end{table*}


\begin{table*}
\begin{center}
\caption{\bf ImageNet ResNet-101} 
\label{table:imagenet_resnet101_config}

\begin{tabular}{ |C{5cm}|| p{0.2cm} |C{5cm}|  }
\hline
\bf{Base Model} &  \multicolumn{2}{c|}{\expertconnectbold} \\ 
\hline
\hline

{CONV: 7$\times$7,64} &  \multirow{5}{*}{\rotatebox[origin=c]{90}{\bf  Stem}} & CONV: 7$\times$7,64 \\ 
{POOL: 3$\times$3,Max,2}  & & {POOL: 3$\times$3,Max,2} \\

\cline{1-1}\cline{3-3} 

{ \Bigg\{ \parbox[]{2.7cm}{CONV: 1$\times$1,64 \\ CONV: 3$\times$3,64 \\ CONV: 1$\times$1,256}  \Bigg\}   $\times3$  } 
& &
{ \Bigg\{ \parbox[]{2.7cm}{CONV: 1$\times$1,64 \\ CONV: 3$\times$3,64 \\ CONV: 1$\times$1,256}  \Bigg\}   $\times3$  } 
\\ 
\cline{1-1}\cline{3-3}

{ \Bigg\{ \parbox[]{2.7cm}{CONV: 1$\times$1,128 \\ CONV: 3$\times$3,128 \\ CONV: 1$\times$1,512}  \Bigg\}   $\times4$  } 
& &
{ \Bigg\{ \parbox[]{2.7cm}{CONV: 1$\times$1,128 \\ CONV: 3$\times$3,128 \\ CONV: 1$\times$1,512}  \Bigg\}   $\times4$  } 
\\ 
\cline{1-1}\cline{3-3}

{ \Bigg\{ \parbox[]{2.7cm}{CONV: 1$\times$1,256 \\ CONV: 3$\times$3,256 \\ CONV: 1$\times$1,1024}  \Bigg\}   $\times23$  } 
& &
{ \Bigg\{ \parbox[]{2.7cm}{CONV: 1$\times$1,256 \\ CONV: 3$\times$3,256 \\ CONV: 1$\times$1,1024}  \Bigg\}   $\times23$  } 
\\ 
\cline{1-1}\cline{3-3}

{ \Bigg\{ \parbox[]{2.7cm}{CONV: 1$\times$1,512 \\ CONV: 3$\times$3,512 \\ CONV: 1$\times$1,2048}  \Bigg\}   $\times3$  } 
& &
{ \Bigg\{ \parbox[]{2.7cm}{CONV: 1$\times$1,512 \\ CONV: 3$\times$3,512 \\ CONV: 1$\times$1,2048}  \Bigg\}   $\times2$  } 
\\ 
\cline{1-1}\cline{3-3}

\hline

{POOL: 7$\times$7,Ave,1} &   \multirow{2}{*}{\rotatebox[origin=c]{90}{\bf Branch}}  & 
{ \Bigg\{ \parbox[]{2.7cm}{CONV: 1$\times$1,512 \\ CONV: 3$\times$3,512 \\ CONV: 1$\times$1,1024}  \Bigg\}   $\times1$ }  \\ 
  &  & {POOL: 7$\times$7,Ave,1}    \\

\hline
{FC: 1000 } 	&           \multicolumn{2}{c|}   {FC\_Gates: 1000 }   \\

\hline
\hline
\multicolumn{3}{|c|}{\bf Data preprocessing} \\
\hline
\parbox[t]{5cm}{Input: $224\times224$\\Image mean subtraction\\Image mirroring} 
& 
\multicolumn{2}{l|}{\parbox{4cm}{Input: $224\times224$\\Image mean subtraction\\Image mirroring}}
\\
\hline
\multicolumn{3}{|c|}{\bf Learning Policy} \\
\hline

\parbox{5.5cm}{
{\footnotesize 
$\bullet$ Learning rate: 0.1, 0.01, 0.001\\ 
$\bullet$ Momentum: 0.9\\
$\bullet$ Weight decay: 0.0001\\ 
$\bullet$ Weight initialization: MSRA~\cite{HeEtAl:ICCV2015} \\
} }
& 
\multicolumn{2}{l|}{\parbox{5cm}{
{\footnotesize 
$\bullet$ Learning rate: 0.1, 0.01, 0.001\\ 
$\bullet$ Gate Learning rate: 10$\times$Learning rate \\
$\bullet$ Momentum: 0.9\\
$\bullet$ Weight decay: 0.0001\\ 
$\bullet$ Weight initialization: MSRA~\cite{HeEtAl:ICCV2015} 
}}}
 
\\
 
\hline 
\end{tabular}

\end{center}
\end{table*}


\begin{table*}[t]
\begin{center}
\caption{\textbf{DICT+2-90k}}
\label{table:synth_config}

\begin{tabular}{ |C{7.5cm}|| p{0.2cm} |C{5.8cm}|  }
\hline
\bf{Base Model} &  \multicolumn{2}{c|}{\expertconnectbold} \\ 
\hline
\hline

{CONV: 5$\times$5,64 } &  \multirow{4}{*}{\rotatebox[origin=c]{90}{\bf  Stem}} & CONV: 5$\times$5,64  \\ 
{POOL: 2$\times$2,Max,2}  & & {POOL: 2$\times$2,Max,2}   \\
\cline{1-1}\cline{3-3}

{CONV: 5$\times$5,128}  & & CONV: 5$\times$5,128  \\ 
{POOL: 2$\times$2,Max,2}  & & {POOL: 2$\times$2,Max,2}   \\
\cline{1-1}\cline{3-3}

{CONV: 3$\times$3,256}  & & CONV: 3$\times$3,256  \\ 
{POOL: 2$\times$2,Max,2}  & & {POOL: 2$\times$2,Max,2}   \\
\cline{1-1}\cline{3-3} 

{CONV: 3$\times$3,512}  & & CONV: 3$\times$3,512  \\ 

\hline

{CONV: 3$\times$3,512}   &  \multirow{3}{*}{\rotatebox[origin=c]{90}{\bf Branch}}  & {CONV: 3$\times$3,512}   \\ 

\cline{1-1}\cline{3-3} 

{FC: 4096 } 	&           &   {FC: 256 }   \\ 
\cline{1-1}\cline{3-3} 

{FC: 4096 } 	&           &   {FC: 256 }   \\ 
\hline

{FC: 90000 } 	&           \multicolumn{2}{c|} {FC\_Gates: 90000 }   \\

\hline
\hline
\multicolumn{3}{|c|}{\bf Data preprocessing} \\
\hline
\parbox[t]{7cm}{Input: $32\times100$\\Image mean subtraction} 
& 
\multicolumn{2}{l|}{\parbox{6.5cm}{Input: $32\times100$\\Image mean subtraction}}
\\
\hline
\multicolumn{3}{|c|}{\bf Learning Policy} \\
\hline
& \multicolumn{2}{l|}{} \\
\parbox{7cm}{
{\footnotesize 
$\bullet$ Incremental learning starting with 5000 classes till
convergence then adding another 5000 till all 90000 are added.\\
$\bullet$ Dropout after fully-connected layers.\\
$\bullet$ Learning rate: 0.01, 0.001, 0.0001\\ 
$\bullet$ Momentum: 0.9\\
$\bullet$ Weight decay: 0.0005\\ 
$\bullet$ Weight initialization: Random\\ 
} }
& 
\multicolumn{2}{l|}{\parbox{7.5cm}{
{\footnotesize 
$\bullet$ Incremental learning starting with 5000 classes till
convergence then adding another 5000 till all 90000 are added.\\
$\bullet$ No Dropout used.\\
$\bullet$ Learning rate: 0.01, 0.001, 0.0001\\ 
$\bullet$ Gate Learning rate: 10$\times$Learning rate \\
$\bullet$ Momentum: 0.9\\
$\bullet$ Weight decay: 0.0005\\
$\bullet$ Weight initialization: Random
}}}
 
\\
 
\hline 
\end{tabular}

\end{center}
\end{table*}

\end{document}